\documentclass[journal, twocolumn]{IEEEtran}
\usepackage[nocompress]{cite}
\usepackage{threeparttable}
\usepackage{graphicx}
\usepackage{stfloats}
\usepackage{amsmath}
\usepackage{amsfonts}
\usepackage{multirow}
\usepackage{multicol}
\usepackage{cite}
\usepackage[numbers]{natbib}
\usepackage{setspace}
\usepackage{color}
\usepackage{makecell}
\usepackage{graphicx,amsmath,booktabs,latexsym}
\usepackage{amssymb}
\usepackage[lined,boxed,ruled]{algorithm2e}
\usepackage{subfigure}
\usepackage[switch]{lineno}
\usepackage{hhline}
\usepackage{amsthm}
\usepackage{cite}
\usepackage{hyperref}
\usepackage{cuted}
\usepackage{float}

\newtheorem{myRemark}{Remark}
\newtheorem{myTheo}{Theorem}
\newtheorem{myAsump}{Assumption}
\newtheorem{myLemma}{Lemma}
\newtheorem{myDef}{Definition}
\newenvironment{myProof}{{\noindent\it Proof}\quad}{\hfill $\square$\par}

\allowdisplaybreaks[4] 

\hypersetup{hidelinks,
	colorlinks=true,
	allcolors=blue,
	pdfstartview=Fit,
	breaklinks=true}

\begin{document}
\begin{sloppypar}
\title{FM3Q: \textcolor{black}{Factorized} Multi-Agent MiniMax Q-Learning for Two-Team Zero-Sum Markov Game
}

\author{
	Guangzheng~Hu,~\IEEEmembership{Graduate Student Member,~IEEE},
	Yuanheng~Zhu,~\IEEEmembership{Senior Member,~IEEE},
	Haoran~Li,~\IEEEmembership{Member,~IEEE},
	and~Dongbin~Zhao,~\IEEEmembership{Fellow,~IEEE}
	
	\thanks{This work was supported in part by the National Key Research and Development Program of China under Grant 2018AAA0102404, in part by the National Natural Science foundation of China under Grant 62293541, and also in part by the Strategic Priority Research Program of Chinese Academy of Sciences under Grant No. XDA27030400.}

	\thanks{G. Hu is with the School of Artificial Intelligence, University of Chinese Academy of Sciences, Beijing 100049, China, and also with the State Key Laboratory of Multimodal Artificial Intelligence Systems, Institute of Automation, Chinese Academy of Sciences, Beijing 100190, China (email : huguangzheng2019@ia.ac.cn).}
	
	\thanks{Y. Zhu, H. Li and D. Zhao are with the State Key Laboratory of Multimodal Artificial Intelligence Systems, Institute of Automation, Chinese Academy of Sciences, Beijing 100190, China, and also with the School of Artificial Intelligence, University of Chinese Academy of Sciences, Beijing 100049, China (email : \{yuanheng.zhu, lihaoran2015, dongbin.zhao\}@ia.ac.cn).}

}

\markboth{IEEE TRANSACTIONS, ~Vol.~,  No.~,  ~2023}%
{Shell \MakeLowercase{\textit{et al.}}: Bare Demo}

\maketitle
%
\begin{abstract}

	Many real-world applications involve some agents that fall into two teams, with payoffs that are equal within the same team but of opposite sign across the opponent team. The so-called two-team zero-sum Markov games (2t0sMGs) can be resolved with reinforcement learning in recent years. However, existing methods are thus inefficient in light of insufficient consideration of intra-team credit assignment, data utilization and computational intractability. In this paper, we propose the individual-global-minimax (IGMM) principle to ensure the coherence between two-team minimax behaviors and the individual greedy behaviors through Q functions in 2t0sMGs. Based on it, we present a novel multi-agent reinforcement learning framework, \textcolor{black}{Factorized} Multi-Agent MiniMax Q-Learning (FM3Q), which can factorize the joint minimax Q function into individual ones and iteratively solve for the IGMM-satisfied minimax Q functions for 2t0sMGs. Moreover, an online learning algorithm with neural networks is proposed to implement FM3Q and obtain the deterministic and decentralized minimax policies for two-team players. A theoretical analysis is provided to prove the convergence of FM3Q. Empirically, we use three environments to evaluate the learning efficiency and final performance of FM3Q and show its superiority on 2t0sMGs.
\end{abstract}


\section{Introduction}\label{sec:introduction}
Markov games (MGs), also known as stochastic games, have been widely used to model the strategic interactions of multiple agents in dynamic environments with multiple states \cite{Markov}. In recent years, multi-agent reinforcement learning (MARL) has achieved impressive success in its application to multi-agent systems \cite{chen2020intelligentcrowd, anastassacos2020partner, zheng2021reward}, particularly in two-player zero-sum games (2p0sMGs) \cite{silver2016mastering, silver2018general, StarCraft, tang2020enhanced}. Some research on 2p0sMGs includes approaches to estimating the value at Nash Equilibrium (NE) based on Bellman-like operators \cite{nashq}. In each optimization process, most of them have to solve the NE of the stage game in each specific state \cite{minimaxq, ffq}.
Whilst it is theoretically possible to solve for NE in 2p0sMGs via linear programming in polynomial time, the optimization over joint-action space suffers from combinatorial explosion as the number of players increases \cite{feng2021neural}. 
One effective approach, population-based multi-agent reinforcement learning (PB-MARL) \cite{malib}, continually generates advanced intelligence by leveraging auto-curricula and has achieved impressive successes in tackling multi-agent tasks, such as the variants of self-play (SP) that are able to achieve expert-level performance on OpenAI Five \cite{OpenAIfive}, and methods based on policy space response oracles (PSRO) that have recently achieved impressive performance on AlphaStar \cite{StarCraft} and Stratego \cite{Pipeline}.  
In addition, the variants of PSRO utilize different parallelized mechanisms and opponent selection to improve the computation and exploration efficiency at learning high-quality best responses \cite{psro,psrorn,Pipeline, mcaleer2021xdo}. However, PB-MARL are relatively data-thirsty because of the intrinsic dynamics arising from multiple agents and populations. In other words, in order to be effective, these algorithms must generate a large amount of new data for training in each curriculum. 

In addition to just two players, many real-world scenarios involve adversarial situations between two teams, which are typically modeled as 2t0sMGs. Intra-team cooperation and inter-team competition both exist, and it has attracted much interest from reinforcement learning, for example when playing complex games like Dota 2 \cite{OpenAIfive}, Starcraft \cite{shao2018starcraft}, and Honor of kings arena \cite{wei2022honor}. 
Of particular interest is the distinction between two lines of research on 2t0sMGs, that only considers inter-team competition and another that also considers intra-team cooperation. 
\textcolor{black}{The former line regards} a team as a single individual and make the 2t0sMGs problem equivalent to 2p0sMGs. A typical example is the success of OpenAIfive \cite{OpenAIfive}, in which the presence of full information sharing between teammates makes it equivalent to 2p0sMGs. When addressing inter-team competition, it employs the SP; however, it does not consider the issue of intra-team cooperation and instead utilized an independent learning. The same applies to Honor of kings arena \cite{wei2022honor}. \textcolor{black}{Another line of research puts} the emphasis on intra-team cooperation to coordinate their strategies \cite{celli2019coordination}, or on the efficiency of computing NE policies \cite{kalogiannis2022efficiently}. These research efforts simplify 2t0sMGs to adversarial team Markov games, in which a team of identically interested players is competing against an adversarial player. However, these methods have yet to address issues such as credit assignment among agents.

In recent years, the development of cooperative MARL methods has been advancing rapidly \cite{yu2021surprising, zhang2022automatic, hu2021event, lai2022multiagent}. 
Several value-decomposition-based algorithms have emerged in the area of addressing credit assignment. These algorithms are based on the Indivadual-Global-Max (IGM) principle, which not only factorizes the joint Q function into the independent ones, but also potentially solves the problem of credit assignment. As far as we know, this method has not yet been applied to 2t0sMGs.

This naturally leads to the following open question we are interested in: Can we extend the IGM condition and design an efficient MARL framework for 2t0sMGs, in consideration of intra-team credit assignment, data utilization and computational intractability? To shed light on this open problem, we aim to enable agents to perform two-team minimax behaviors in a decentralized way, and learn such behaviors in an online mode. In this paper, we combine game theory, fitted Q-iteration (FQI), and factorized multi-agent Q-learning to reach the goal. We summarize our contributions as follows:
\begin{itemize}
	\item We define a novel Individual-Global-MiniMax (IGMM) principle for 2t0sMGs, which specifies the coherence between two-team minimax behaviors and the individual greedy behaviors through Q functions. 

	\item With the IGMM principle, \textcolor{black}{Factorized} Multi-Agent MiniMax Q-Learning (FM3Q), which can factorize the joint minimax Q function into individual ones and synchronously optimize the policies of all agents in two teams, is proposed. Moreover, an online learning algorithm with neural networks is proposed to implement FM3Q and learn the two-team minimax policies.

	\item On the basis of FQI, we prove the global convergence of FM3Q. Empirically, we evaluate the online FM3Q algorithm and baselines on Pong, MPE, and RoboMaster, and demonstrate the outstanding performance of the FM3Q.
\end{itemize}

\section{Related Work}\label{sec:relatedWork}
\subsection{MARL Research on 2t0sMGs}

A lot of studies treat a team as a single individual, and 2t0sMGs as 2p0sMGs. 2p0sMGs has been widely recognized as the benchmark setting for MARL. The Minimax-Q algorithm focuses on the zero-sum setting with asymptotic convergent guarantees \cite{minimaxq}. In the same vein as Minimax-Q learning, asymptotic convergence has also been established for other Q-learning variants beyond the zero-sum setting with coordination among agents, such as Nash-Q \cite{nashq}, Friend-or-Foe Q-Learning \cite{ffq}, and M2QN \cite{zhu2020online}. 
Each agent in the above algorithms solves a linear program or quadratic program to solve a matrix game at each iteration. As the number of players increases, the optimization over joint-action space suffers from combinatorial explosion. 

On the other hand, PB-MARL-type algorithms integrate reinforcement learning with dynamical population selection methods to produce auto-curricula. In PSRO, each player finds an approximate best response to its opponents’ meta-strategies, and the new policies are added into policy sets for the next iteration \cite{psro}. $\text{PSRO}_{\text{rN}}$ uses rectified Nash mixtures, that is, each learner only plays against other learners that it already beats, to encourage policy diversity \cite{psrorn}.
Alpha-PSRO introduces the alpha-rank multi-agent evaluation metric and preference-based best response in PSRO and shows promising performance in computing equilibria \cite{alpharank}. There is an increasing emphasis on reducing the intensive computation of PSRO, and many new methods have been proposed. Deep Cognitive Hierarchies (DCH) is one such approach, which parallelizes PSRO to enable scaling to larger games and improve learning speed \cite{psro}. 
The scalability of PSRO can be enhanced through the implementation of a hierarchical pipeline of reinforcement learning agents, where each agent at a higher level is trained against the policies generated by agents at lower levels in the hierarchy \cite{Pipeline}. 
\textcolor{black}{Neural Extensive-Form Double Oracle (NXDO) can be viewed as a version of PSRO where the restricted game allows mixing population policies not only at the root of the game, but at every infostate \cite{mcaleer2021xdo}.}
Nevertheless, PB-MARL approaches still have certain limitations, including (1) the inefficient utilization of data from previous iterations, as it is discarded and not used to learn new policies, and (2) the need for additional computational resources for meta-solving and empirical payoff evaluation.

In addition, other researchers are dedicated to studying intra-team cooperation and other problems. 
Soft Team Actor-Critic (STAC) \cite{celli2019coordination} is proposed to make intra-team members associate shared meanings to signals that are initially uninformative. Independent Policy GradientMax (IPGMAX) \cite{anagnostides2023algorithms} is able to compute stationary $\epsilon$-approximate NE in adversarial team Markov games with computational complexity that is polynomial in all the natural parameters of the game. They simplify 2t0sMGs to adversarial team Markov games, in which a team of players is competing against an adversarial player, not a team.
\textcolor{black}{Adversarial collaborative learning
(ACL) exploits friend-or-foe Q-learning and mean-field theory,  trains the friends and oppenents via adversarial
max and min steps, and suffers from the curse of dimensionality \cite{luo2020multiagent}. 
However, ACL treats 2t0sMGs as 2p0sMGs and ignores credit assignment among agents.}
Additionally, empirical policy optimization (EPO) propose a novel multi-player reinforcement learning method, in which
the parameters are trained based on the whole history of experience \cite{zhu2022empirical}, but EPO is only applicable to the case that any party in the game is a single agent. 

\subsection{Value Factorization and Fitted Q-Iteration}

Value decomposition has been increasingly popular in centralized traning and decentralized execution (CTDE)-based MARL for cooperative tasks. The IGM principle \cite{qmix} of equivalence between joint greedy action and individual greedy action is critical. Existing value-decomposition algorithms satisfy the IGM consistency by expressing the joint Q function with global state as a function of individual Q functions with local observation. Additivity and monotonicity are respectively considered in VDN \cite{vdn} and QMIX \cite{qmix}, with some variants proposed to relax the monotonicity constraint of QMIX \cite{wqmix, chai2021unmas}. These methods have achieved impressive performance on cooperative MARL but there are still some issues to be addressed in applying them to
competitive tasks.

Deep Q-learning, as one of the core components of DRL, has shown great success in solving complicated problems. Fitted Q-Iteration (FQI) is based on iterative Bellman error minimization and utilizes a specific Q-function class to iteratively optimize the empirical Bellman error on a dataset \cite{fqi,cfqi}. FQI has unique advantages from a theoretical perspective, and there is a growing trend to adopt FQI for theoretical and empirical analysis in DRL \cite{efqi}. A closely related recent paper, Factorzed Multi-Agent Fitted Q-Iteration (FMA-FQI) \cite{fmafqi} models the iterative training procedure of multi-agent Q-learning using empirical Bellman error minimization and formally analyzes cooperative MARL with value factorization. 
Therefore, if these methods can be applied to the competitive MARL, it would also be conducive to theoretical analysis.

\section{Preliminaries}\label{sec:Background}

To facilitate theoretical analysis, we divide two teams into the Protagonists (Pro) and the Antagonists (Ant), and use the Decentralized Partial-Observation Markov Decision Process (Dec-POMDP) framework to model the decision-making problem in 2t0sMGs.

Dec-POMDP in 2t0sMGs is described as a tuple $\mathcal{MG} := \left\langle \mathcal{S}, \boldsymbol{\mathcal{A}},  \boldsymbol{\mathcal{B}}, \boldsymbol{\mathcal{X}}, \boldsymbol{\mathcal{Y}}, \Lambda, \Omega, \mathcal{N}, \mathcal{M}, P, {R}, \gamma \right\rangle$,  where 
$\mathcal{S}$ denotes the state space;
$\mathcal{N} \equiv \{ 1, ..., n\}$ and $\mathcal{M} \equiv \{ 1, ..., m\}$ are the finite sets of agents, and $\boldsymbol{\mathcal{X}} = \left\{ \mathcal{X}_i \right\}_{i=1,2,...,n}$ 
and $\boldsymbol{\mathcal{Y}} = \left\{ \mathcal{Y}_i \right\}_{i=1,2,...,m}$
represents the sets of observations of each agent in Pro and Ant, respectively; 
$\Lambda_i(s): \mathcal{S} \rightarrow \mathcal{X}_i$ 
and $\Omega_i(s): \mathcal{S} \rightarrow \mathcal{Y}_i$ 
are the observation function that determines the private observation, and the agent $i$ receives a private observation by $x_i = \Lambda_i(s)$ or $y_i = \Omega_i(s)$.
$\boldsymbol{\mathcal{A}} = \left\{ \mathcal{A}_i \right\}_{i=1,2,...,n}$ 
and $\boldsymbol{\mathcal{B}} = \left\{ \mathcal{B}_i \right\}_{i=1,2,...,m}$ denote the sets of actions of agents in Pro and Ant.
$P(s'|s, \boldsymbol{a}, \boldsymbol{b}) : \mathcal{S} \times \boldsymbol{\mathcal{A}}  \times \boldsymbol{\mathcal{B}} \times \mathcal{S} \rightarrow [0,1]$ represents the state transition function, where $\boldsymbol{a}=(a_1,...,a_n)$ and $\boldsymbol{b}=(b_1,...,b_m)$ are the joint action.
${R} : \mathcal{S} \times \boldsymbol{\mathcal{A}} \times \boldsymbol{\mathcal{B}} \rightarrow \mathbb{R}$  indicate the set of reward function of Pro.
$\gamma \in [0,1]$ is the discount factor.

\textcolor{black}{In many real-world 2t0sMGs tasks, data sampling is a significantly challenging endeavor. Therefore, the primary concern shifts towards obtaining high-performance models using less data rather than acquiring accurate Nash equilibrium solutions at an unconstrained sampling cost. Hence, in this paper, our emphasis is on addressing issues related to intra-team credit assignment, data utilization, and computational intractability in 2t0sMGs to attain superior performance while working within the confines of deterministic policy settings.}
Each agent in Pro has an observation-action history $\tau_i \in \mathcal{T}_i \equiv(\mathcal{X}_i \times \mathcal{A}_i)^*$, on which it conditions a deterministic policy $\pi_i\left(a_i \mid \tau_i\right): \mathcal{T}_i  \times \mathcal{A}_i \rightarrow\{0,1\}$, and aims to maximize the expected discounted return $\mathbb{E}\left[ \sum_{t=0}^\infty \gamma^t r_{t} \right]$, where $r_{t} \sim {R}(s_t, \boldsymbol{a}_t, \boldsymbol{b}_t)$. Each agent in Ant has an observation-action history $v_i \in \mathcal{V}_i \equiv(\mathcal{Y}_i \times \mathcal{B}_i)^*$, on which it conditions a stochastic policy $\mu_i\left(b_i \mid v_i\right): \mathcal{V}_i \times \mathcal{B}_i \rightarrow\{0,1\}$, and aims to minimize the above-mentioned discounted return.
$\boldsymbol{\pi}=(\pi_1,...,\pi_n)$ and $\boldsymbol{\mu}=(\mu_1,...,\mu_m)$ are the joint policy, and $\boldsymbol{\tau}=(\tau_1,...,\tau_n)$ and $\boldsymbol{v}=(v_1,...,v_m)$ represent the joint observation-action history.
We overload $\tilde{s}= \left\langle \boldsymbol{\tau},  \boldsymbol{v}, s \right\rangle$ for simplicity. From a centralized perspective, $\left(\boldsymbol{\pi}, \boldsymbol{\mu}\right)$ has a joint minimax Q function $Q_{\mathrm{tot}}^{\boldsymbol{\pi}, \boldsymbol{\mu}}(\tilde{s}, \boldsymbol{a}, \boldsymbol{b})=\mathbb{E}_{[\tilde{s}, \boldsymbol{a}, \boldsymbol{b}]_{t+1: \infty}}\left[\sum_{k=0}^{\infty} \gamma^k r_{t+k}\right]$, and a joint minimax V value function $V_{Q_{\mathrm{tot}}}^{\boldsymbol{\pi}, \boldsymbol{\mu}}(\tilde{s})=\mathbb{E}_{[\tilde{s}]_{t: \infty}}\left[\sum_{k=0}^{\infty} \gamma^k r_{t+k}\right]$.  Our goal is to find the decentralized and deterministic policies $\boldsymbol{\pi}^*$ and $\boldsymbol{\mu}^*$
, which satisfy $V_{Q_{\mathrm{tot}}}^{\boldsymbol{\pi}^*, \boldsymbol{\mu}^*}(\tilde{s})=\min\limits_{\boldsymbol{\mu}}{\max\limits_{\boldsymbol{\pi}}}{V_{Q_{\mathrm{tot}}}^{\boldsymbol{\pi},\boldsymbol{\mu}}(\tilde{s})}$. 
Further, we show that $\min\limits_{\boldsymbol{\mu}}{\max\limits_{\boldsymbol{\pi}}}{V_{Q_{\mathrm{tot}}}^{\boldsymbol{\pi},\boldsymbol{\mu}}(\tilde{s})} = \min\limits_{\boldsymbol{b}} \max\limits_{\boldsymbol{a}}Q_{\text {tot}}^{\boldsymbol{\pi}^*, \boldsymbol{\mu}^*}(\tilde{s}, \boldsymbol{a}, \boldsymbol{b})$ under deterministic policies. 
Thus, we transform the goal to obtain $Q_{\text {tot}}^{\boldsymbol{\pi}^*, \boldsymbol{\mu}^*}(\tilde{s}, \boldsymbol{a}, \boldsymbol{b})$ and perform a global arg min max on it.
In what follows, we drop the superscript $\boldsymbol{\pi}, \boldsymbol{\mu}$ for simplicity, and we define the superb Q function $Q_{\text {tot}}^{*}$ and let $Q_{\text {tot}}^{*}(\tilde{s}, \boldsymbol{a}, \boldsymbol{b}) =Q_{\text {tot}}^{\boldsymbol{\pi}^*, \boldsymbol{\mu}^*}(\tilde{s}, \boldsymbol{a}, \boldsymbol{b})$.

\section{Factorized Multi-Agent MiniMax Fitted Q-Iteration} \label{sec:fm3q}

In this section, we first introduce the IGMM principle to ensure the coherence between two-team minimax behaviors and individual greedy behaviors through Q functions. Then, based on the IGMM principle, we propose the FM3Q framework to factorize the joint minimax Q function into individual ones.  FM3Q draws inspiration from the FQI \cite{fqi} and iteratively optimizes the policies of all agents synchronously.
Finally, we prove the convergence of FM3Q.

\vspace{-1.5mm}
\subsection{Individual-Global-MiniMax}\label{sec:igmm}

One idea for directly solving the 2t0sMGs is to divide all agents into two groups, with one group maximizing the rewards and the other group minimizing the rewards, using the joint minimax Q function and constructing linear programming to solve for the NE. However, the optimization over joint-action space suffers from combinatorial explosion as the number of agents increases. Under certain conditions, can we factorize the joint minimax Q function into the individual ones in order to reduce the computational complexity? We need to establish coherence between two-team minimax behaviors and individual greedy behaviors through Q functions.
Inspired by the IGM condition used in cooperative MARL, we propose the Indivadual-Global-MiniMax (IGMM) principle to enforce the consistency of action selection between the global joint minimax Q function, denoted as $Q_{\mathrm{tot}}$, and the individual Q functions, denoted as $[Q_i^+ ]_{i=1}^{n}$ and $[Q_i^- ]_{i=1}^{m}$ for Pro and Ant, respectively. The IGMM principle can be represented as follows:
	\begin{equation} \label{minimax_tot}
	\begin{aligned}
		& \arg\min\limits_{\boldsymbol{b}}{\max\limits_{\boldsymbol{a}}{Q_{\mathrm{tot}}(\tilde{s}, \boldsymbol{a}, \boldsymbol{b})}} 
		=\arg\max\limits_{\boldsymbol{a}}{\min\limits_{\boldsymbol{b}}{Q_{\mathrm{tot}}(\tilde{s}, \boldsymbol{a}, \boldsymbol{b})}}  \\
		& = \left(
		\arg\max\limits_{a_1}{Q_1^+},
		\ldots,
		\arg\max\limits_{b_m}{Q_m^-}  \right)
	\end{aligned}
\end{equation}
Coherence can be achieved when the greedy decentralized policies are determined by an arg max over the $[Q_i^+ ]_{i=1}^{n}$ and $[Q_i^- ]_{i=1}^{m}$, and a global arg min max performed on $Q_{\mathrm{tot}}$ yields the same result as a set of individual arg max operations performed on $[Q_i^+ ]_{i=1}^{n}$ and $[Q_i^- ]_{i=1}^{m}$. Based on the IGMM, multi-agent reinforcement learning algorithms satisfying the CTDE paradigm can be designed for 2t0sMGs, which not only meet the requirements of distributed execution in many real-world tasks but also significantly reduce computational complexity and improve the efficiency of algorithm training.

To ensure the rationality of the IGMM, it is necessary to impose constraints on it.
\textcolor{black}{We define $f\left(\left[Q_i^+ \left( \tau_i, a_i \right)\right]_{i=1}^{n},  \left[Q_j^- \left( v_j, b_j \right)\right]_{j=1}^{m}, s \right)  = Q_{\mathrm{tot}}(\tilde{s}, \boldsymbol{a}, \boldsymbol{b})$, and the constraints on the relationship between $Q_{\mathrm{tot}}$ and each $Q_i^+$ and $Q_j^-$ as two monotonicities}:
$\frac{\partial Q_{tot}}{\partial Q_i^+} \geq 0, \forall i \in \mathcal{N}$ and $\frac{\partial Q_{tot}}{\partial Q_j^-} \leq 0, \forall j \in \mathcal{M}$, which are adequate to ensure the coherence referred to above. This is demonstrated by the following theorem.

\begin{myTheo} \label{igmm_tot}
	If $\frac{\partial Q_{\mathrm{tot}}}{\partial Q_i^+} \geq 0, \forall i \in \mathcal{N}$ and $\frac{\partial Q_{\mathrm{tot}}}{ \textcolor{black}{ \partial Q_j^-}} \leq 0, \forall j \in \mathcal{M}$, then the IGMM principle can be satisfied.
\end{myTheo}
\begin{myProof}
	Since $\frac{\partial Q_{\mathrm{tot}}}{\partial Q_i^+} \geq 0, \forall i \in \mathcal{N}$, the following holds for any $\boldsymbol{a}$ and $Q_{\mathrm{tot}}$:
	\begin{footnotesize}
	\begin{align*}
		&f\left([Q_i^+ \left( \tau_i, a_i \right)]_{i=1}^{n},  [Q_j^- \left( v_j, b_j \right)]_{j=1}^{m}, s \right)    \\
		&\leq f\left(\max _{a_1} Q_1^+\left(\tau_1, a_1\right), \ldots, Q_n^+\left(\tau_n, a_n\right),  [Q_j^- \left( v_j, b_j \right)]_{j=1}^{m}, s \right)   \\
		&\leq f\left( \left[\max _{a_i} Q_i^+ \left( \tau_i, a_i\right) \right]_{i=1}^{n} , [Q_j^- \left( v_j, b_j \right)]_{j=1}^{m}, s \right)  
	\end{align*}
    \end{footnotesize}
   \noindent \textcolor{black}{This indicates that the joint Q-value increases with the individual Q-values of Pro.} Similarly, since $\frac{\partial Q_{\mathrm{tot}}}{\partial Q_j^-} \leq 0, \forall j \in \mathcal{M}$, the following holds for any $\boldsymbol{b}$ and $Q_{\mathrm{tot}}$:
	\begin{footnotesize}
	\begin{align*}
		&f\left([Q_i^+ \left( \tau_i, a_i \right)]_{i=1}^{n}, [Q_j^- \left( v_j, b_j \right)]_{j=1}^{m}, s\right)   \\
		&\geq  f\left([Q_i^+ \left( \tau_i, a_i \right)]_{i=1}^{n}, \max _{b_1} Q_1^-\left(v_1, b_1\right), \ldots, Q_m^-\left(v_m, b_m\right), s \right)   \\
		&\geq  f\left([Q_i^+ \left( \tau_i, a_i \right)]_{i=1}^{n},  \left[\max _{b_j} Q_j^- \left( v_j, b_j \right) \right]_{j=1}^{m}  , s\right)  
	\end{align*}
	\end{footnotesize} 
\textcolor{black}{This indicates that the joint Q-value decreases with the individual Q-values of Ant.}
    \textcolor{black}{We show that $\max \limits_{\boldsymbol{a}} f\left(\cdot \right) = f\left( \left[\max _{a_i} Q_i^+ \right]_{i=1}^{n} , [Q_j^- ]_{j=1}^{m}, s \right) $ and $\min \limits_{\boldsymbol{b}} f\left(\cdot\right) = f\left([Q_i^+]_{i=1}^{n},  \left[\max _{b_j} Q_j^- \right]_{j=1}^{m}  , s\right).$ 
    Thus, 
	\begin{align*}
		&\max _{\boldsymbol{a}} \min _{\boldsymbol{b}} Q_{\mathrm{tot}}(\tilde{s}, \boldsymbol{a}, \boldsymbol{b}) \\
		&:= \max _{\boldsymbol{a}} \min _{\boldsymbol{b}} f \left([Q_i^+ ]_{i=1}^{n}, [Q_j^- ]_{j=1}^{m} , s\right) \\
		& = f\left( \max_{a_1} Q_1^+, \ldots, \max _{b_m} Q_m^- , s\right).
	\end{align*}
\noindent $\min  \limits_{\boldsymbol{b}} \max  \limits_{\boldsymbol{a}} Q_{\mathrm{tot}}(\tilde{s}, \boldsymbol{a}, \boldsymbol{b}) = f\left( \max \limits_{a_1} Q_1^+, \ldots, \max \limits_{b_m} Q_m^- , s\right)$ represents a similar process.} 
Letting 
	$\left(a_1^*, \ldots, a_n^*, b_1^*, \ldots, b_m^* \right)=\left(
		\arg\max\limits_{a_1}{Q_1^+},
		\ldots,
		\arg\max\limits_{b_m}{Q_m^-}  \right) $, we have that 
	\begin{align*}
		&f\left([Q_i^+ \left( \tau_i, a_i^* \right)]_{i=1}^{n}, [Q_j^- \left( v_j, b_j^* \right)]_{j=1}^{m}, s \right) \\
		& = f\left( \max _{a_1} Q_1^+\left(\tau_1, a_1\right), \ldots, \max _{b_m} Q_m^- \left(v_m, b_m\right),s \right) \\
		& =\max _{\boldsymbol{a}} \min _{\boldsymbol{b}} Q_{\mathrm{tot}}(\tilde{s}, \boldsymbol{a}, \boldsymbol{b})=\min _{\boldsymbol{b}} \max _{\boldsymbol{a}} Q_{\mathrm{tot}}(\tilde{s}, \boldsymbol{a}, \boldsymbol{b})
	\end{align*}
	Hence, 	$$\begin{aligned}
		&\quad \arg\min\limits_{\boldsymbol{b}}{\max\limits_{\boldsymbol{a}}{Q_{\mathrm{tot}}(\tilde{s}, \boldsymbol{a}, \boldsymbol{b})}} 
		=\arg\max\limits_{\boldsymbol{a}}{\min\limits_{\boldsymbol{b}}{Q_{\mathrm{tot}}(\tilde{s}, \boldsymbol{a}, \boldsymbol{b})}}  \\
		& = \left(
		\arg\max\limits_{a_1}{Q_1^+},
		\ldots,
		\arg\max\limits_{b_m}{Q_m^-}  \right)
	\end{aligned}$$
\end{myProof}

\vspace{-2mm}

\subsection{Factorized Multi-Agent MiniMax Q-Learning}\label{sec:fm3q_}

Recall that our goal is to obtain the superb Q function $Q_{\text {tot}}^{*}$ and perform a global arg min max on it. Directly solving it faces the problems of high computational complexity and may not satisfy the requirement of distributed execution in some environments.
The IGMM condition ensures consistency of action selection between the global joint minimax Q function and the individual Q functions. 
Therefore, based on it, we propose a new multi-agent method framework called Factorized Multi-Agent MiniMax Q-Learning (FM3Q) for solving 2t0sMGs. FM3Q factorizes the joint minimax-Q function subject to the IGMM principle and optimizes the policies through the use of factorized individual Q-functions, which satisfies the CTDE paradigm. To facilitate theoretical analysis and clarify our approach, we introduce the framework of Fitted Q-Iteration (FQI) \cite{fqi}, to optimize the policies of all agents by iteratively minimizing the empirical minimax Bellman error.
We overload $\tilde{Q}= \left\langle Q_{\mathrm{tot}},  [Q_i^+ ]_{i=1}^{n}, [Q_j^- ]_{j=1}^{m} \right\rangle$ to indicate their association to facilitate further discussions.
Additionally, we greedily turn $Q_{\mathrm{tot}}(\tilde{s}, \boldsymbol{a}, \boldsymbol{b}) \in \mathbb{R}^{\boldsymbol{\mathcal{T}} \times \boldsymbol{\mathcal{V}} \times {\mathcal{S}} \times \boldsymbol{\mathcal{A}} \times \boldsymbol{\mathcal{B}} }$ into \textcolor{black}{$V_{Q_{\mathrm{tot}}}(\tilde{s}):=\min _{\boldsymbol{b}} \max _{\boldsymbol{a}}   Q_{\text {tot}}\left(\tilde{s}, \boldsymbol{a}, \boldsymbol{b} \right)     \in \mathbb{R}^{\boldsymbol{\mathcal{T}} \times \boldsymbol{\mathcal{V}}\times {\mathcal{S}}}  $ }to facilitate below discussions.

FM3Q is an iterative optimization framework like FQI that is based on a provided dataset $D$. Let $T$ denote the total number of iterations . We first randomly initialize $\tilde{Q}^{(0)}$ at time $0$ from $\mathcal{Q}^{\text {IGMM }}$, as defined in Definition \ref{def_1}.
In the subsequent training, $\tilde{Q}$ is iteratively updated at iteration $t \in [0, T-1]$ as:

\noindent
\begin{small}
	\begin{equation} \label{operator_new}
		\begin{aligned}
			&\tilde{Q}^{(t+1)} \leftarrow \mathcal{T}_D^{\mathrm{IGMM}} \tilde{Q}^{(t)} \equiv \\
			& \underset{\tilde{Q} \in \mathcal{Q}^{\mathrm{IGMM}}}{\arg \min } \underset{(\tilde{s}, \boldsymbol{a}, \boldsymbol{b}, r, \tilde{s}') \sim D}{\mathbb{E}} \textcolor{black}{ \left[
			e^{(t)} \left(\tilde{s}, \boldsymbol{a}, \boldsymbol{b}, r, \tilde{s}' \right) 
			- Q_{\mathrm{tot}}(\tilde{s}, \boldsymbol{a}, \boldsymbol{b})\right]}^2
		\end{aligned}
	\end{equation} 
\end{small}
where \begin{equation} \label{td_target}
	\begin{aligned}
		& e^{(t)} \left(\tilde{s}, \boldsymbol{a}, \boldsymbol{b}, r, \tilde{s}' \right)=r+\gamma \underset{\tilde{s}'} {\mathbb{E}} \left[\min _{\boldsymbol{b}^{\prime}} \max _{\boldsymbol{a}^{\prime}}   Q^{(t)}_{\text {tot}}\left(\tilde{s}', \boldsymbol{a}^{\prime}, \boldsymbol{b}^{\prime}\right)\right]
	\end{aligned}
\end{equation} 
denotes the one-step temporal difference target at time $t$. Based on $\tilde{Q}^{(t+1)}$, we can construct decentralized policies by individual value functions as:
\begin{center}
	$\forall i \in \mathcal{N}, \pi^{(t+1)}_i \left(\tau_i\right)=\underset{a_i \in \mathcal{A}_i}{\arg \max } {Q_i^{+}}^{(t+1)}\left(\tau_i, a_i\right)$; \\
	$\forall j \in \mathcal{M}, \mu^{(t+1)}_j \left(v_j\right)=\underset{b_j \in \mathcal{B}_i}{\arg \max } {Q_j^{-}}^{(t+1)}\left(v_j, b_j\right)$.
\end{center}

\begin{myAsump}
	(Exploratory Data Collection). The dataset $D$ is collected by two joint exploratory policies $\boldsymbol{\pi}^D$ and $\boldsymbol{\mu}^D$ satisfying $
	\forall(\boldsymbol{\tau} \times \boldsymbol{v} \times \boldsymbol{a} \times \boldsymbol{b} ) \in \boldsymbol{\mathcal{T}} \times \boldsymbol{\mathcal{V}} \times \boldsymbol{\mathcal{A}} \times \boldsymbol{\mathcal{B}}, \boldsymbol{\pi}^D(\boldsymbol{a} \mid \boldsymbol{\tau})>0$ and $\boldsymbol{\mu}^D(\boldsymbol{b} \mid \boldsymbol{v})>0$.
\end{myAsump}
\begin{myDef} \label{def_1}
	
	FM3Q specifies the $\tilde{Q}$ function class with a complete IGMM principle realization
	\begin{small}
	\begin{align*} 
		&\mathcal{Q}^{\mathrm{IGMM}}=\left\{\tilde{Q} \mid \right. \\
		&\left. \arg{\min\limits_{\boldsymbol{b}}\max\limits_{\boldsymbol{a}}{Q_{\mathrm{tot}}(\tilde{s}, \boldsymbol{a}, \boldsymbol{b})}} = \left(	
		\begin{array}{c}
			\arg\max\limits_{a_1}{Q_1^+(\tau_1, a_1)}\\
			\vdots \\
			\arg\max\limits_{b_m}{Q_m^-(v_m, b_m)}
		\end{array}
		\right), \right. \\
		& \left.   Q_{\mathrm{tot}}(\tilde{s}, \boldsymbol{a}, \boldsymbol{b}) \in \mathbb{R}^{|\boldsymbol{\mathcal{T}} \times \boldsymbol{\mathcal{V}} \times \boldsymbol{\mathcal{A}} \times \boldsymbol{\mathcal{B}} \times\mathcal{S}| }, \left[Q_i^+\right]_{i=1}^n \in \mathbb{R}^{|\mathcal{T} \times \mathcal{A}|^n},  \right. \\
		& \left.   \left[Q_j^-\right]_{j=1}^m \in \mathbb{R}^{|\mathcal{V} \times \mathcal{B}|^m} \right\}
	\end{align*}
    \end{small}
\end{myDef}

The FM3Q algorithm minimizes the empirical minimax Bellman error by utilizing global reward signals obtained from the Pro. 
After iterative updates using FM3Q, the final two-team minimax Q function and the individual Q functions can be obtained.
With the individual Q functions, the greedy action selection can be searched in the individual action spaces $\mathcal{A}_i$ or $\mathcal{B}_j$, rather than the joint action space $\boldsymbol{\mathcal{A}} \times \boldsymbol{\mathcal{B}}$, which significantly reduces the computation costs when distributedly executing.

\vspace{-2mm}

\subsection{Convergence of FM3Q}\label{sec:convergence}
\begin{myLemma} \label{lemma_emm}
	The empirical minimax Bellman operator	$\mathcal{T}_D^{\mathrm{IGMM}}$ in Equation (\ref{operator_new}) is a $\gamma$-contraction.
\end{myLemma}

\begin{myProof}	
	First, for a given $\tilde{Q}= \left\langle Q_{\mathrm{tot}},  [Q_i^+ ]_{i=1}^{n}, [Q_j^- ]_{j=1}^{m} \right\rangle$, the analytical solution of $\mathcal{T}_D^{\mathrm{IGMM}} \tilde{Q}$ is presented as $\tilde{q} = \left\langle q_{\mathrm{tot}},  [q_i^+ ]_{i=1}^{n}, [q_j^- ]_{j=1}^{m}\right\rangle$.
	For a sample $(\tilde{s}, \boldsymbol{a}, \boldsymbol{b}, r, \tilde{s}')$ in the dataset $D$, let 
	\begin{equation*}
		\begin{split}
			&\boldsymbol{a}^{*}, \boldsymbol{b}^{*}=\left[a_i^{*}\right]_{i=1}^n, \left[b_j^{*}\right]_{b=1}^m =\arg \min _{\mathbf{b} \in \boldsymbol{\mathcal{B}} } \max _{\boldsymbol{a} \in \boldsymbol{\mathcal{A}}}  e(\tilde{s}, \boldsymbol{a}, \boldsymbol{b}, r, \tilde{s}').
		\end{split}	
	\end{equation*}
   We show that 
   \begin{equation}
   	\label{qt}
   	q_{\mathrm{tot}} \left(\tilde{s}, \boldsymbol{a}, \boldsymbol{b} \right) = e \left(\tilde{s}, \boldsymbol{a}, \boldsymbol{b}, r, \tilde{s}' \right)
   \end{equation}
   and  its corresponding decentralized local Q functions $\left[q_i^{+}\right]_{i=1}^n,\left[q_j^{-}\right]_{j=1}^m$:
   \begin{equation}
   	\label{q+}
   	q_i^+\left(\tau_i, a_i\right)= \begin{cases}1, & \text { when } a_i=a_i^{*} \\ 0, & \text { when } a_i \neq a_i^{*}\end{cases}
   \end{equation}
   and
   \begin{equation}
   	\label{q-}
   	q_j^-\left(v_j, b_j\right)= \begin{cases}1, & \text { when } b_j=b_j^{*} \\ 0, & \text { when } b_j \neq b_j^{*}\end{cases}.
   \end{equation}
   It is easy to see that $ \left\langle q_{\mathrm{tot}},  [q_i^+ ]_{i=1}^{n}, [q_j^- ]_{j=1}^{m}\right\rangle$ satisfies the IGMM principle, and the empirical minimax Bellman error reach the minimum zero: $\sum p_D(\tilde{s}, \boldsymbol{a}, \boldsymbol{b}, r, \tilde{s}') \left(e \left(\tilde{s}, \boldsymbol{a}, \boldsymbol{b}, r, \tilde{s}'\right)
   - q_{\mathrm{tot}}(\tilde{s}, \boldsymbol{a}, \boldsymbol{b})\right)^2=0$, 
    where $p_D(\tilde{s}, \boldsymbol{a}, \boldsymbol{b}, r, \tilde{s}')$ represents the probability of a sample in the dataset $D$. This proves that $\tilde{q}$ is the analytical solution of $\mathcal{T}_D^{\mathrm{IGMM}} \tilde{Q}$.

	Second, let us show that for all $\tilde{s}$, $\left|V_{Q_{\mathrm{tot}}}(\tilde{s}) - V_{Q_{\mathrm{tot}}^{\prime}}(\tilde{s})\right| \leq \min _{\boldsymbol{b}} \max _{\boldsymbol{a}}  \left|Q_{\mathrm{tot}}(\tilde{s}, \boldsymbol{a}, \boldsymbol{b})-Q_{\mathrm{tot}}^{\prime}(\tilde{s}, \boldsymbol{a}, \boldsymbol{b})\right|$. Then, we can derive the following:
	\textcolor{black}{\begin{small}$$	
	\begin{aligned} 
		& \min _{\boldsymbol{b}} \max _{\boldsymbol{a}} Q_{\text {tot}} -  \min _{\boldsymbol{b}}\max _{\boldsymbol{a}}  Q_{\text {tot}}^{\prime}
		= - \left[ \min _{\boldsymbol{b}} \max _{\boldsymbol{a}} Q_{\text {tot}}^{\prime} -  \min _{\boldsymbol{b}}\max _{\boldsymbol{a}}  Q_{\text {tot}} \right]   \\
		&  \leq -\min _{\boldsymbol{b}} \left[ \max _{\boldsymbol{a}} Q_{\text {tot}}^{\prime}- \max _{\boldsymbol{a}} Q_{\text {tot}} \right] 
		= \max _{\boldsymbol{b}} \left[ \max _{\boldsymbol{a}} Q_{\text {tot}} - \max _{\boldsymbol{a}} Q_{\text {tot}}^{\prime} \right] \\
		& \leq \max _{\boldsymbol{b}} \max _{\boldsymbol{a}} \left[ Q_{\text {tot}}- Q_{\text {tot}}^{\prime} \right]  \leq \max _{\boldsymbol{b}} \max _{\boldsymbol{a}} \left| Q_{\text {tot}} - Q_{\text {tot}}^{\prime} \right|
	\end{aligned}$$ \end{small}}
	\hspace*{-0.2cm}
	\parbox{0\textwidth}{
	\textcolor{black}{\begin{small}$$\begin{aligned}  
		&  \min _{\boldsymbol{b}} \max _{\boldsymbol{a}} Q_{\text {tot}} -  \min _{\boldsymbol{b}} \max _{\boldsymbol{a}} Q_{\text {tot}}^{\prime} \geq \min _{\boldsymbol{b}} \left[ \max _{\boldsymbol{a}} Q_{\text {tot}} - \max _{\boldsymbol{a}} Q_{\text {tot}}^{\prime} \right] \\
		& \geq \min _{\boldsymbol{b}} \left[ -\max _{\boldsymbol{a}} \left(Q_{\text {tot}}^{\prime} -  Q_{\text {tot}}\right) \right]  \geq -\max _{\boldsymbol{b}} \max _{\boldsymbol{a}} \left[ Q_{\text {tot}}^{\prime}- Q_{\text {tot}} \right]\\
		&  \geq -\max _{\boldsymbol{b}} \max _{\boldsymbol{a}} \left| Q_{\text {tot}}^{\prime} - Q_{\text {tot}} \right| 
	\end{aligned}$$ \end{small}}}

	Therefore, we can conclude:$\left|V_{Q_{\mathrm{tot}}}(\tilde{s}) - V_{Q_{\mathrm{tot}}^{\prime}}(\tilde{s})\right| \leq \max _{\boldsymbol{b}} \max _{\boldsymbol{a}}  \left|Q_{\mathrm{tot}}(\tilde{s}, \boldsymbol{a}, \boldsymbol{b})-Q_{\mathrm{tot}}^{\prime}(\tilde{s}, \boldsymbol{a}, \boldsymbol{b})\right|$.
	Using this,
	\begin{footnotesize}
	$$
	\begin{aligned}
		&\left\| \left(\mathcal{T}_D^{\mathrm{IGMM}} Q\right)_{\mathrm{tot}} - \left(\mathcal{T}_D^{\mathrm{IGMM}} Q^{\prime}\right)_{\mathrm{tot}} \right\|_{\infty} 
		= \left\|    q_{\mathrm{tot}} - q^{'}_{\mathrm{tot}}  \right\|_{\infty}  \\
		&= \gamma\left\|\mathbb{E}\left[V_{Q_{\mathrm{tot}}}\right]-\mathbb{E}\left[V_{Q_{\mathrm{tot}}^{\prime}}\right]\right\|_{\infty} \leq \gamma\left\|V_{Q_{\mathrm{tot}}}-V_{Q_{\mathrm{tot}}^{\prime}}\right\|_{\infty} \\
		&=\gamma \max _{\boldsymbol{\tau}} \max _{\boldsymbol{v}} \max _{\tilde{s}} \left|V_{Q_{\mathrm{tot}}}(\tilde{s})-V_{Q_{\mathrm{tot}}^{\prime}}(\tilde{s})\right| \\
		& \leq \gamma \max _{\boldsymbol{\tau}} \max _{\boldsymbol{v}}   \max _{\tilde{s}} \max _{\boldsymbol{a}} \max _{\boldsymbol{b}}  \left|Q_{\mathrm{tot}}(\tilde{s}, \boldsymbol{a}, \boldsymbol{b})-Q_{\mathrm{tot}}^{\prime}(\tilde{s}, \boldsymbol{a}, \boldsymbol{b})\right| \\
		&=\gamma\left\|Q_{\mathrm{tot}}-Q_{\mathrm{tot}}^{\prime}\right\|_{\infty}
	\end{aligned}
	$$	
	\end{footnotesize}
	
	Now, we can prove that $\mathcal{T}_D^{\mathrm{IGMM}}$ is a $\gamma$-contraction.
\end{myProof}

\begin{myTheo} \label{convergency}
	FM3Q globally converges to the superb Q function under deterministic setting in the \textcolor{black}{Dec-POMDPs} if IGMM condition is satisfied. 
\end{myTheo}
\begin{myProof}
	Recall that \begin{small}
	\noindent $Q_{\text {tot}}^*(\tilde{s}, \boldsymbol{a}, \boldsymbol{b})=\max\limits_{\boldsymbol{\pi} \in \boldsymbol{\Pi}} \min\limits_{\boldsymbol{\mu} \in \boldsymbol{\Phi}} Q_{\text {tot}}^{\boldsymbol{\pi}, \boldsymbol{\mu}}(\boldsymbol{\tau},\boldsymbol{v}, s, \boldsymbol{a}, \boldsymbol{b})$\end{small}, where $Q_{\text {tot}}^*$ 
	denotes the superb Q function, and $\boldsymbol{\Pi}$ and $\boldsymbol{\Phi}$ are the space of all deterministic policies of both sides if IGMM condition is satisfied.


	First, we want to prove that  $Q^*_{\text {tot }}=\left(\mathcal{T}_D^{\text {IGMM}} Q^*\right)_{\text {tot }}$.
	
   \begin{small}
	$$
	\begin{aligned} 
		&Q_{\text {tot}}^*(\tilde{s}, \boldsymbol{a}, \boldsymbol{b})=\max _{\boldsymbol{\pi}} \min _{\boldsymbol{\mu} } Q_{\text {tot}}^{\boldsymbol{\pi}, \boldsymbol{\mu}}(\tilde{s}, \boldsymbol{a}, \boldsymbol{b})\\
		&=\max _{\boldsymbol{\pi}} \min _{\boldsymbol{\mu} } \left\{ r+\gamma \mathbb{E}\left[V_{Q_\mathrm{tot}}^{\boldsymbol{\pi}, \boldsymbol{\mu}}\left(\tilde{s}'\right)\right]\right\} \\ 
		&= r+\gamma \mathbb{E}\left[\max _{\boldsymbol{\pi}} \min _{\boldsymbol{\mu} }V_{Q_\mathrm{tot}}^{\boldsymbol{\pi}, \boldsymbol{\mu}}\left(\tilde{s}'\right)\right] \\ 
		&= r+\gamma \mathbb{E}\left[\max _{\boldsymbol{\pi}} \min _{\boldsymbol{\mu} }Q_{\text {tot}}^{\boldsymbol{\pi}, \boldsymbol{\mu}}(\tilde{s}', \boldsymbol{\pi}(\boldsymbol{\tau}^{\prime}), \boldsymbol{\mu}(\boldsymbol{v}^{\prime}) )\right] \\ 
		&= \textcolor{black}{r+\gamma \mathbb{E}\left[\max _{\boldsymbol{\pi},\boldsymbol{a}^{\prime}} \min _{\boldsymbol{\mu},\boldsymbol{b}^{\prime} }Q_{\text {tot}}^{\boldsymbol{\pi}, \boldsymbol{\mu}}(\tilde{s}', \boldsymbol{a}', \boldsymbol{b}'))\right]} \\ 
		&= r+\gamma \mathbb{E}\left[\max _{\boldsymbol{a}^{\prime}} \min _{\boldsymbol{b}^{\prime} }Q_{\text {tot}}^*(\tilde{s}', \boldsymbol{a}', \boldsymbol{b}')\right] \\ 
		&= r+\gamma \mathbb{E}\left[V_{Q_{\text {tot}}^*}\left(\tilde{s}'\right)\right] = \left(\mathcal{T}_D^{\text {IGMM}} Q^*\right)_{\text {tot }}
	\end{aligned}
	$$
	\end{small}
	
	Second, $\forall Q_{\mathrm{tot}}^{\prime} \in \mathcal{Q}^{\mathrm{IGMM}}$,
	\begin{small}
	$$
	\begin{aligned}
		&\left\|Q_{\text {tot }}^*-\left(\mathcal{T}_D^{\text {IGMM}} Q^{\prime}\right)_{\text {tot }}\right\|_{\infty} \\ &=\left\|\left(\mathcal{T}_D^{\text {IGMM}} Q^*\right)_{\text {tot }}-\left(\mathcal{T}_D^{\text {IGMM}} Q^{\prime}\right)_{\text {tot }}\right\|_{\infty}  \leq \gamma\left\|Q_{\text {tot }}^*-Q_{\text {tot }}^{\prime}\right\|_{\infty} .
	\end{aligned}
	$$ 
	\end{small}
	Thus, FM3Q will globally converge to a superb Q function.
\end{myProof}

\section{Online Learning of FM3Q} \label{sec:fm3qonline}


In this section, we give a formal introduction to the online learning of FM3Q, which includes the processes of forward propagation, action selection, and training. Moreover, we provide a novel coordinator when training that significantly differ from previous DRL training methods. It is of importance for training FM3Q, and failure to adhere to it may lead to suboptimal outcomes. The architecture and the detailed process of the sampling and training of FM3Q.

\begin{figure*}[!ht]
	
	\begin{center}
	
		\includegraphics[width=0.9\linewidth]{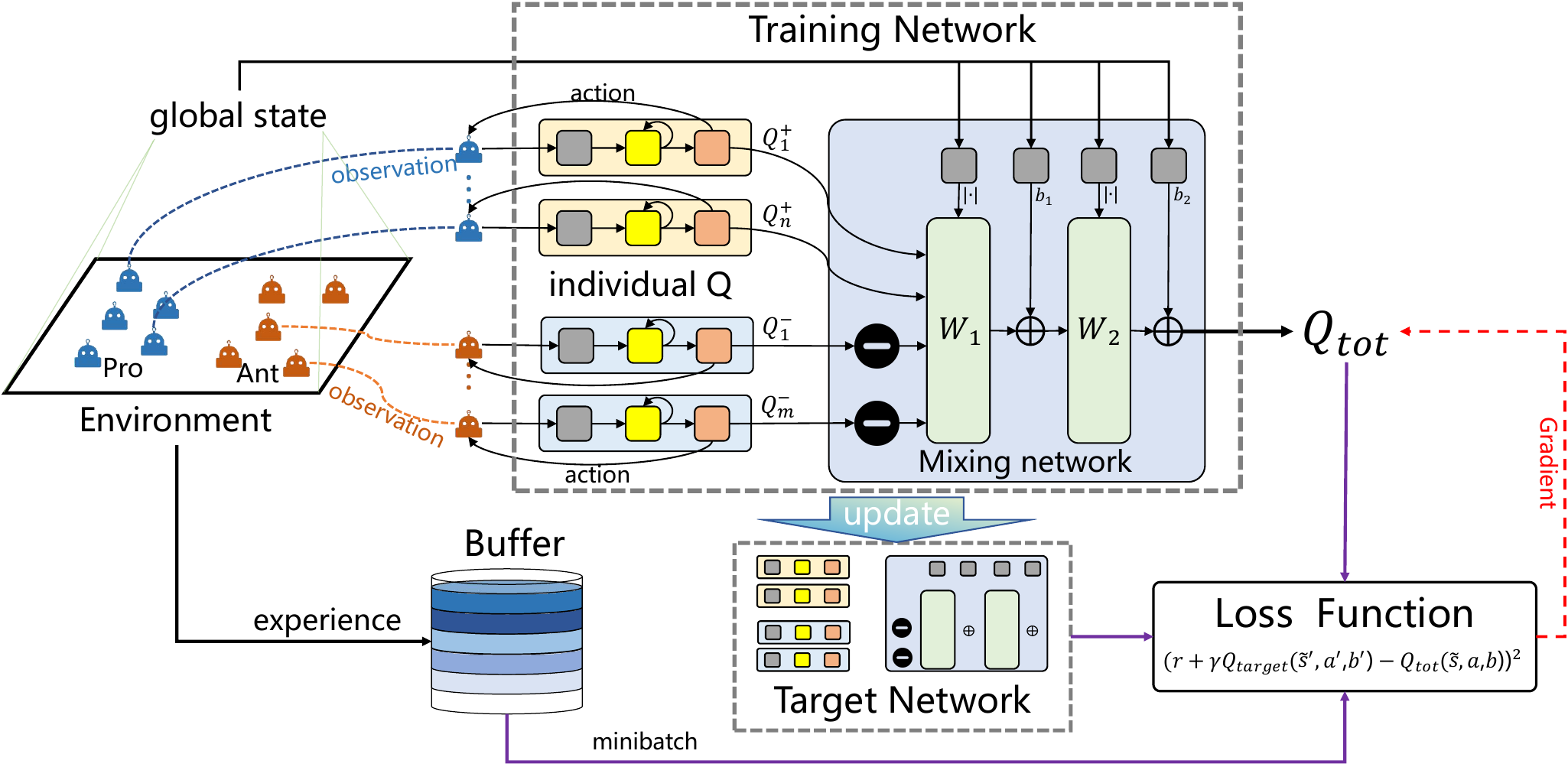}
		
		\caption{\textcolor{black}{The overall architecture of FM3Q, where two-team agents acquire observations from the environment and derive their actions and corresponding individual Q-values through their utility modules. The individual Q-values of the Ant undergo the negation process and, combined with the individual Q-values of the Pro, are processed through a mixing network to yield their minimax joint Q-values. The black arrows indicate that FM3Q interacts with the environment and stores experience when performing execution. The purple arrows indicate the foward propagation process when training model. The red arrow show the gradient flow.
		 }}
		
		\label{fig:struts}
		
	\end{center}
\end{figure*}

\subsection{Forward Propagation and Action Selection}

The overall process of forward propagation is carried out in the direction indicated by the black arrowheads in the left part of Figure \ref{fig:struts}. The current observation of the agent ($x_i$ or $y_j$) is processed through its respective Q networks, as shown in the pink boxes, and the action ($a_i$ or $b_j$) and the Q function ($Q_i^+(\tau_i, a_i)$ or $Q_j^-(v_i, b_i)$) are determined through action selection. The Q function of the Pro agents are directly input into the mixing network, while the ones of the Ant agents first undergo a negative module (as shown in the blue box). Finally, the mixing network outputs the joint minimax Q value. The specific implementation details are represented below.

For each agent $i$ in Pro, there is an individual Q-network $Q_i^{+}\left(\tau_i, a_i;\theta_i^+ \right)$, parameterized by $\theta_i^+$, similarly in Ant, Q-network $Q_i^{-}\left(v_i, b_i;\theta_i^- \right)$, parameterized by $\theta_i^-$. Neural networks in these modules can switch flexibly with the application scenario. For example, the Gated Recurrent Unit (GRU), as a memory-based method, can be selected in the partially observable scenario for better performance in execution. 
Set $\epsilon$ value to guide whether it is exploration or exploitation. The goal of $\epsilon$-greedy exploration is to make all possible actions under a certain state have a non-zero probability of being selected for execution. We can obtain the policy for agent $i$ in Pro by $\pi_i \left(\cdot | \tau_{i} \right) = \mathcal{G}\left( Q_i^{+}\left(\tau_{i}, \cdot | \theta^+ \right), \epsilon\right) $, detailed as
{
\begin{equation}
	\small
\pi_i \left(a_i | \tau_{i} \right) =
\begin{cases}\frac{\epsilon}{|\mathcal{A}|}+1-\epsilon & \text { if } a_i =\underset{a \in \mathcal{A}}{\operatorname{argmax}} Q_i^{+}\left(\tau_{i}, a | \theta^+ \right) \\ \frac{\epsilon}{|\mathcal{A}|} & \text { otherwise }\end{cases}
\end{equation}}
The Pro agent selects the action by $a_{i} \sim \pi_{i}( \cdot | \tau_{i})$. The same holds true for $ \mu_{j}( b_{j} | v_{j})$ and  $Q_j^{-}\left(v_{j}, b | \theta^- \right)$ in the Ant agents.

To ensure that $\frac{\partial Q_{tot}}{\partial Q_i^+} \geq 0, \forall i \in \mathcal{N}$ and $\frac{\partial Q_{tot}}{\partial Q_j^-} \leq 0, \forall j \in \mathcal{M}$ as required by IGMM condition, we need to construct a monotonic mixing network to mix the corresponding individual Q functions.  
We formulate the monotonic mixing network as $Q_{\mathrm{tot}} = \text{Mix} \left( [Q_i^+ ]_{i=1}^{n}, -[Q_j^- ]_{j=1}^{m}, s; \phi\right)$, parameterized by $\phi$, and let it satisfy $\frac{\partial Q_{tot}}{\partial Q_i^+} \geq 0, \forall i \in \mathcal{N}$ and $\frac{\partial Q_{tot}}{\partial Q_j^-} \leq 0, \forall j \in \mathcal{M}$, respectively, \textcolor{black}{by constraining the weights within the mixing network to a range greater than 0}. 
The mixing network takes the Pro network outputs and the negative of the Ant network outputs as input and mixes them monotonically, producing the values of $Q_{\mathrm{tot}}$. The weights of the mixing network $\omega$, as shown in Figure \ref{fig:struts}, are generated by individual hypernetworks, each of which receives the state variable $s$ as input and produces the weights for a specific layer of the mixing network. The absolute activation function enforces the non-negativity of the mixing weights in each hypernetwork. Biases are also generated by hypernetworks, using the same approach, but without the non-negativity constraint. The introduction of the global state $s$ is aimed at allocating different weights to different agents, potentially addressing the issue of credit assignment. Benefiting from the monotonicity constraint in Eq. (\ref{minimax_tot}), minimaximizing joint $Q_{\mathrm{tot}}$ is the equivalent of maximizing individual Q of all agents, resulting in and allowing for superb individual action to maintain consistency with minimax joint action. Thus, a centralized factorizable $Q_{\mathrm{tot}}$ parameterized by $\boldsymbol{\theta}^+, \boldsymbol{\theta}^-,\phi$ is used to
estimate the joint minimax Q function as follows:
{
\begin{equation} \label{qtot}
	\small
	\begin{aligned}
		&\quad Q_{\mathrm{tot}}(\tilde{s}, \boldsymbol{a}, \boldsymbol{b}; \boldsymbol{\theta}^+, \boldsymbol{\theta}^-,\phi) \\ 
		&=\text{Mix} \left( \left[Q_i^+\left(\tau_i, a_i;\theta_i^+ \right)\right]_{i=1}^{n}, -\left[Q_j^-\left(v_j, b_j;\theta_i^- \right)\right]_{j=1}^{m}, s; \phi\right)
	\end{aligned}
\end{equation}
}

\vspace{-8mm}
\subsection{Loss and Training}

As shown in the right part of Figure \ref{fig:struts}, the agents interact with the environment, and the generated data is stored in the replay buffer, which differs from the ones used in prior methods as described in Remark \ref{rmk_buffer}. Samples from the buffer are then used to calculate the loss, with the coordinator, as described in Remark \ref{coordinator}, controlling the update frequency of the target network to ensure optimization stability. Finally, the parameters of the network are updated via gradient backpropagation in the direction indicated by the red arrow in Figure \ref{fig:struts}. The entire online learning process of FM3Q is illustrated in Algorithm \ref{alg:algrithm2}. The details are indicated below.

\begin{algorithm}[t]
	\caption{Online learning of FM3Q.}
	\label{alg:algrithm2}
	\LinesNumbered
	
	Initialize the network parameters, the replay buffer $D$, the current buffer size $L$, the batch size $B$, and the update times after a sampling $U$.
	
	\KwIn{the number of episodes $M$}
	
	\KwOut{$[\pi_i ]_{i=1}^{n}$, $[\mu_j ]_{j=1}^{m}$}
	
	\For{ $episode = 1 \ to \ M$ }
	{
		\While{not terminal}
		{

			Compute the individual Q function 
			
			Sample the action for each agent
			
			Interact with the environment
		}
	
		{Store the trajectory in the replay buffer $D$}

		\textcolor{black}{{Sample $U$ batch of samples with size $B=L/U$ }  \\
		\For{ $count = 1 \ to \ U$ }
		{
		    Update the traning networks by minimizing the loss Equation (\ref{loss}) based on the batch
	    }
		Update the target networks}

	}

\end{algorithm}

Based on Equation (\ref{td_target}), the TD target is substituted with approximate target value, using parameters $\hat{\boldsymbol{\theta}}^+, \hat{\boldsymbol{\theta}}^-,\hat{\phi}$ from some previous iteration to let the target network provide a stable supervisory signal and avoid the training instability: 
\begin{equation} \label{qtarget}
\begin{aligned} 
	& e=r+\gamma Q_{\text {tot }}\left(\tilde{s}', \boldsymbol{a}', \boldsymbol{b}'; \hat{\boldsymbol{\theta}}^+, \hat{\boldsymbol{\theta}}^-,\hat{\phi}\right)
\end{aligned}
\end{equation}

The whole network trains the parameters
$\boldsymbol{\theta}^+, \boldsymbol{\theta}^-,\phi$ by back-propagation of the loss $\mathcal{L}$. The loss calculation is shown as follows:
\begin{equation} \label{loss}
	\mathcal{L} = \textcolor{black}{\underset{(\tilde{s}, \boldsymbol{a}, \boldsymbol{b}, r, \tilde{s}') \sim D}{\mathbb{E}} \left( e - Q_{\mathrm{tot}}(\tilde{s}, \boldsymbol{a}, \boldsymbol{b}; \boldsymbol{\theta}^+, \boldsymbol{\theta}^-,\phi)\right)^2}
\end{equation}

\begin{myRemark} 
	The replay buffer enables the use of historical data for training, which greatly enhances sample efficiency. Furthermore, \textcolor{black}{randomly sampling from the replay buffer breaks the correlation between data samples, reducing biases introduced by consecutive sampling. This helps reduce inter-sample correlations and potential sample selection biases, preventing learning instability due to dependencies among the data}. In competitive multi-agent tasks, the game space is typically both transitive and cyclic \cite{psrorn}. Specifically, when the replay buffer covers a small space, the minimax Q solution will be limited in the game space covered by the current data. Therefore, we recommend using the largest replay buffer possible when training FM3Q and, in some cases, incorporating all data generated by interaction. We present experimental evidence in the experiments (Section \ref{sec:ab}) that validates the efficiency of this approach through ablation studies.
	\label{rmk_buffer}
\end{myRemark}
\begin{myRemark}
	Because Q-based methods regard predicting the Q value as a regression problem, which requires a supervisory signal at each update. DQN \cite{mnih2015human} introduces the target network, which is only updated with the training network parameters every $U$ steps and is held fixed between individual updates. As above mentioned, in FM3Q, the distribution of the data in the replay buffer is widely divergent from that generated by the current policy. In order to let the target network provide a more stable supervisory signal, we suggest incorporating as much data as possible in the training process before updating the target network. Of course, this suggestion can be moderately relaxed according to different environments. 
	The amount of data in the replay buffer keeps increasing with training.
	\textcolor{black}{We draw inspiration from the EPO \cite{zhu2022empirical} and modified the update approach for the target network in FM3Q. We denote the current size of the replay buffer as $L$ and the batch size as $B$. Recall that we hope as much data as possible participates in a round of training before updating the target network. It is recommended that ($B=L/U$), and following the update of the training network, the target network should be updated directly, which is presented in line 10 of Algorithm \ref{alg:algrithm2}.}
	\label{coordinator}
\end{myRemark}

\section {Experiments}\label{sec:Experiments}
\begin{figure}[!t]
	\begin{center}
		
		\subfigure[Wimblepong 2v2]{
			\label{fig:pong}
			\includegraphics[width=0.28\linewidth, height=0.215\linewidth]{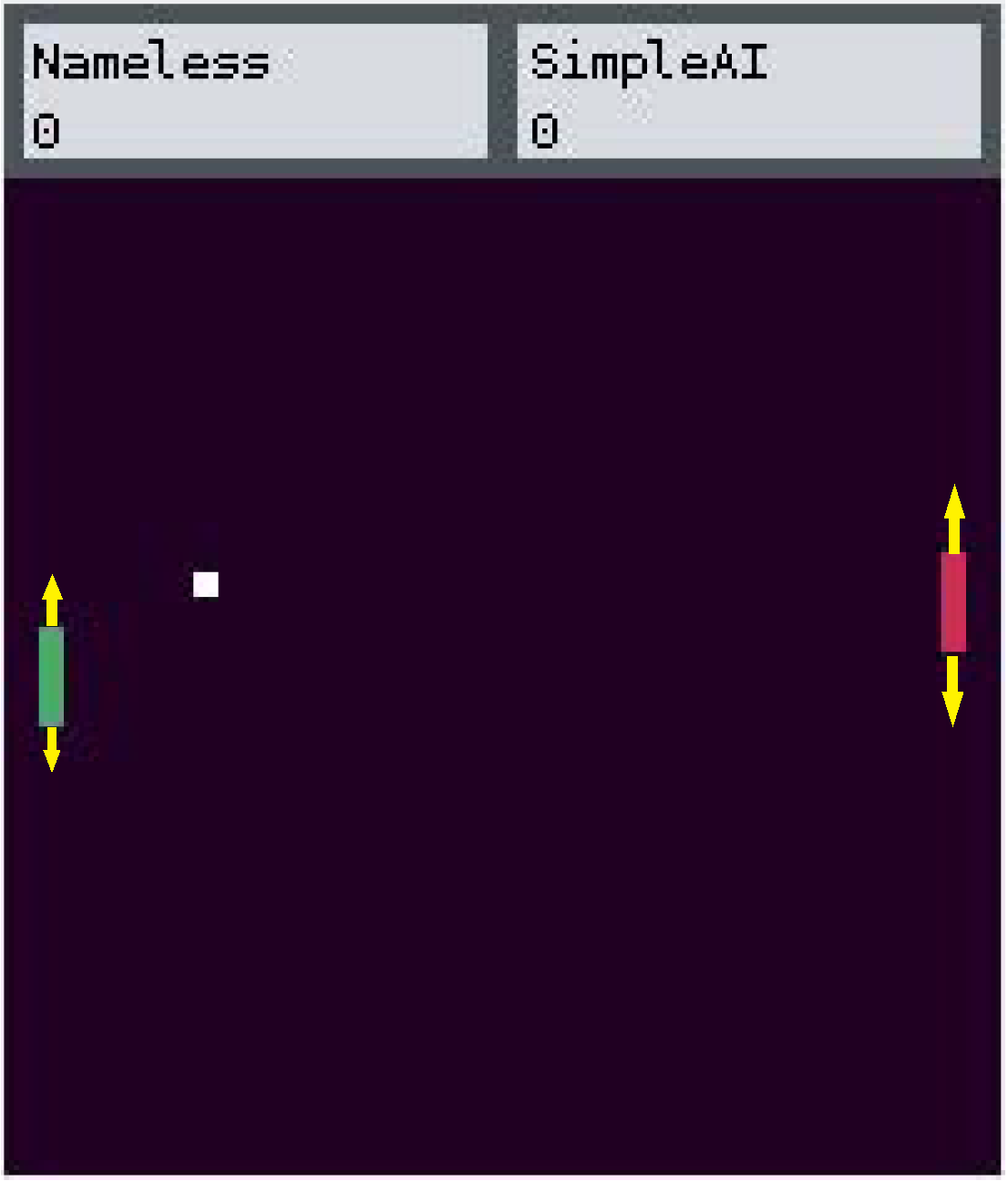}
		}
		\subfigure[MPE 3v3]{
			\label{fig:mpe}
			\includegraphics[width=0.28\linewidth, height=0.215\linewidth]{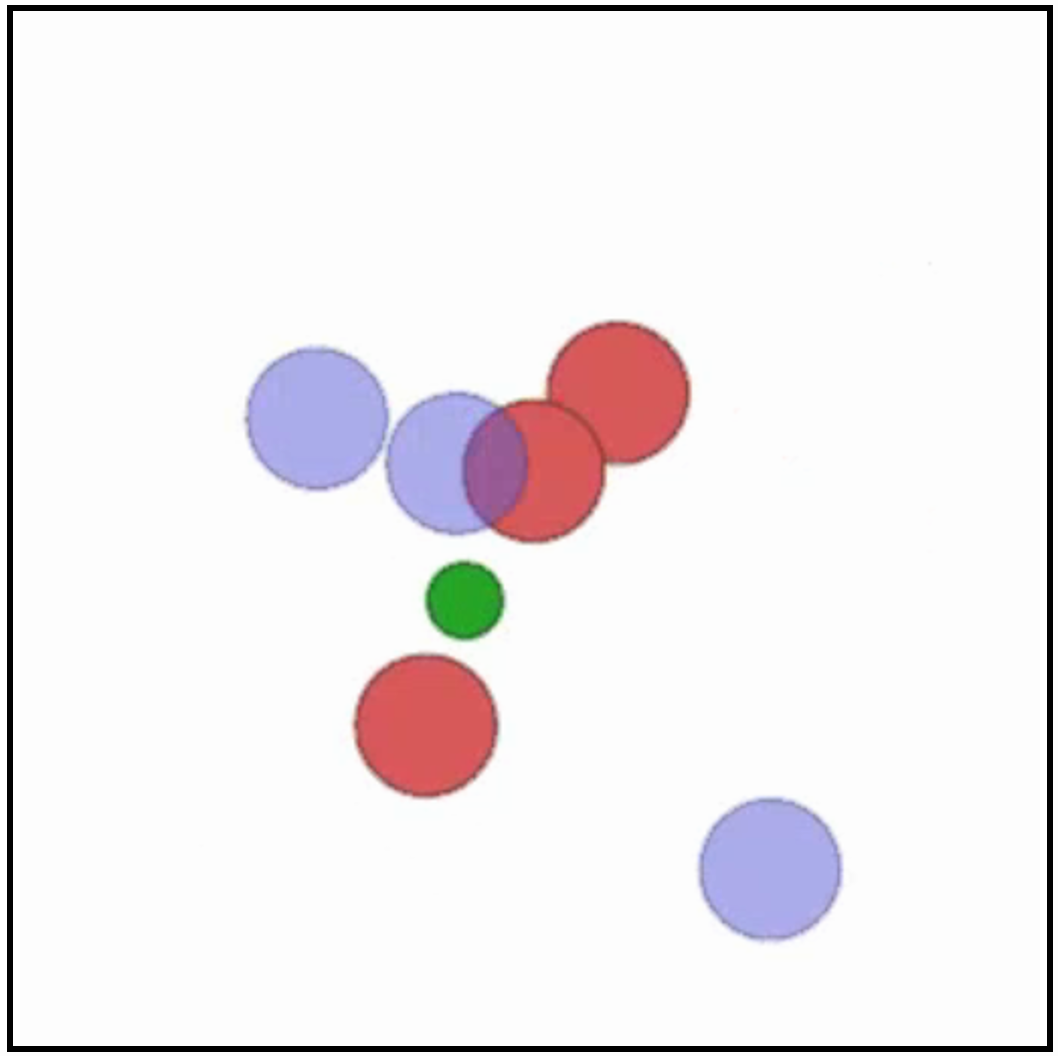}
		}
		\subfigure[RoboMaster 2v2 ]{
			\label{fig:rm}
			\includegraphics[width=0.30\linewidth]{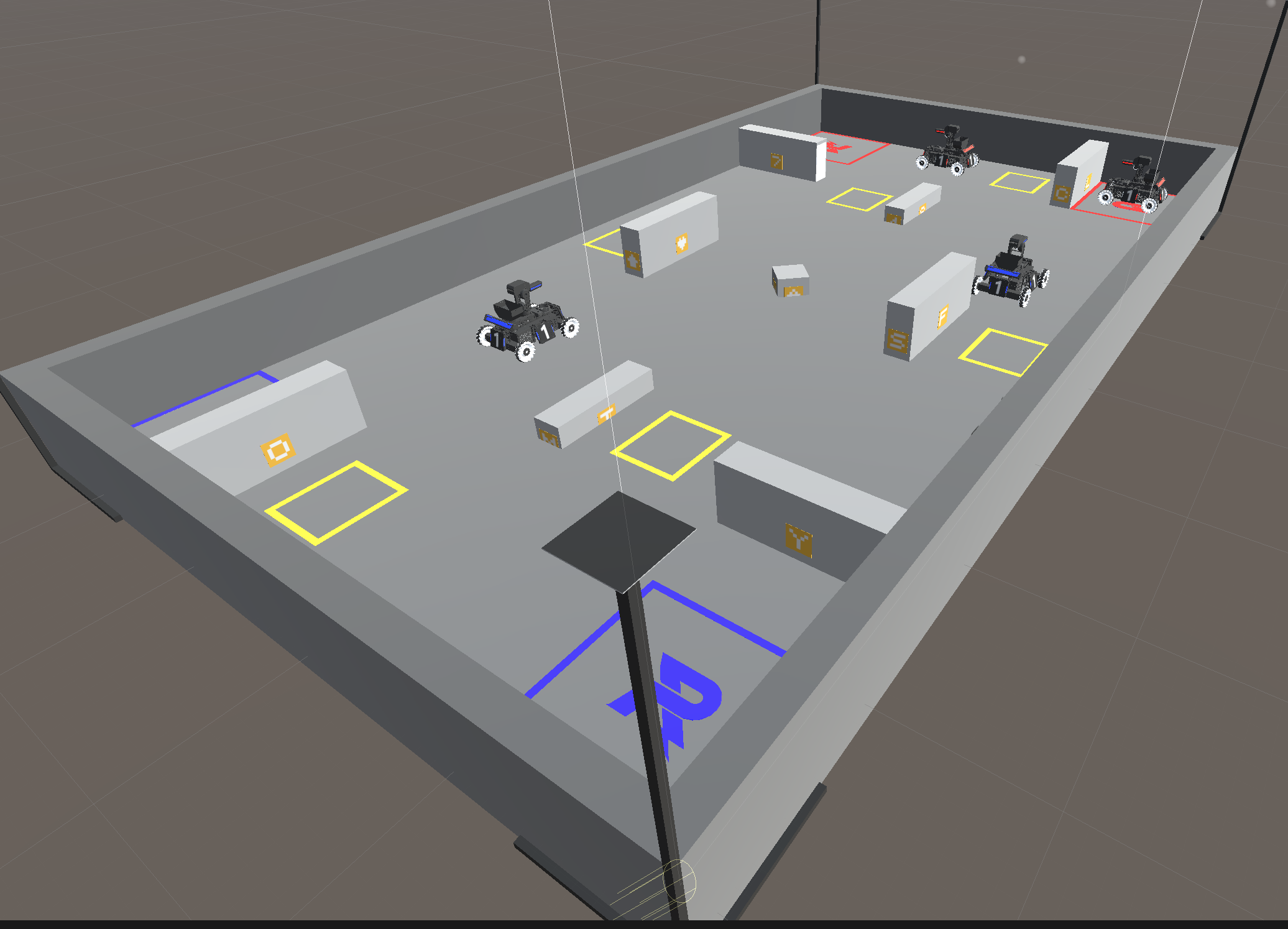}
		}
		\caption{\textcolor{black}{Illustrations of Wimblepong, MPE, and RoboMaster. (a) A green and a red paddle represent two players, respectively, and a white circle represents a ball. (b) Three blue agents and three red agents move, collide, and obstruct each other to maneuver their teams as close as possible to a green target. (c) There are two blue robots, two red robots, and nine obstacles.}}
		
		\label{fig:envs}
		
	\end{center}
\end{figure}

To verify the effectiveness of FM3Q and its superiority over other methods, we conduct experimental validation in three scenarios. Wimblepong is shown in Figure \ref{fig:pong}, MPE is shown in Figure \ref{fig:mpe}, and RoboMaster, which simulates competition between two groups of robots in the real world, is illustrated in Figure \ref{fig:rm}. 

\noindent
\textbf{Wimblepong 2v2} \cite{zhu2022empirical}. \textcolor{black}{It is based on a 2-player version of the Atari game Pong. We modify the environment to create a cooperative-competitive setting involving two teams of two agents each. In this setup, the outputs of the two agents within each team are combined and applied to a paddle as a collective force.
In discrete action settings, the agent takes one of three actions: moving up or down, or staying in place. The game state consists of the positions of the two paddles and the position and velocity of the ball. If a player misses a ball, it receives a -10 reward, and the opponent receives a +10 reward.}
\\
\textbf{MPE 3v3} \cite{terry2021pettingzoo}.  A simple multi-agent particle world with a continuous observation and discrete action space, along with some basic simulated physics. Each of the two sides, red and blue, consists of three agents that collaborate with their allies and compete with the other group to seize the green target. The group with the closest average distance to the target receives a higher reward. Each agent can take one of five actions: moving up or down, moving left or right, or staying in place.
\\
\textbf{RoboMaster 2v2} \cite{NeuronsMAE}. We introduce a multi-robot environment that involves both intra-team cooperation and inter-team competition, named RoboMaster. Robots cooperate with teammates and fight opponents by shooting. The robots defeat the opponents as much as possible within a certain time limit to win the game. The state consists of information about itself, its ally, and two opponents, including position, angle, candidate points, HP, bullet count, and the time remaining. The robot determines which candidate point goes (4-discrete) and which opponent is hit (2-discrete). This environment adopts three reward functions: hitpoint damage dealt ($r_h = 0.02$), enemy units killed ($r_k = 3$), and a bonus for winning the battle ($r_w = 20$).

\textcolor{black}{
We consider SP \cite{OpenAIfive}, PSRO \cite{psro}, EPO \cite{zhu2022empirical}, and NXDO \cite{mcaleer2021xdo} to be popular methods that can be employed to address 2t0sMGs. 
SP involves an agent improving its performance through repeated interaction with itself, while PSRO pertains to a game-solving framework that iteratively refines strategies by considering response dynamics in the space of policies.
EPO introduces an innovative multiplayer reinforcement learning approach where parameters are trained using the whole historical experience and optimized through proximal policy gradient.
Unlike PSRO, which mixes the best responses at the root node of the game, NXDO mixes the best responses at each infostate.
We run all experiments with FM3Q and all baselines, and each experiment is repeated eight times to reduce the impact of randomness. For the sake of fairness, we utilize QMIX as an individual in the policy populations of SP, PSRO, and NXDO and set some identical hyperparameters, such as network and learning rate. All methods related to FM3Q adhere to the recommendations mentioned in Remark \ref{coordinator}. Some important hyperparameters used for training are shown in Appendix.
In each scenario, the five methods save the learned models at different phases of their training for subsequent performance testing and comparison. We evaluate the performance of the five methods from the following aspects: In Section \ref{sec:bot}, we compare the performance differences between the models obtained by three algorithms during the training process and the script-based bots. In Section \ref{sec:baseline}, we compare the exploitability of each method against the others and their approximate NashConv during the training process. In Section \ref{sec:process}, we investigate the optimization trends of FM3Q by comparing the performance of models at different phases during training. In Section \ref{sec:ab}, we validate the impact of buffer size on the training of FM3Q through ablation.}

\begin{figure*}[!ht]
	
	\begin{center}
		\setcounter{subfigure}{0}
		\subfigure[Wimblepong]{
			\label{fig:pong_bot}
			\includegraphics[width=0.3\linewidth]{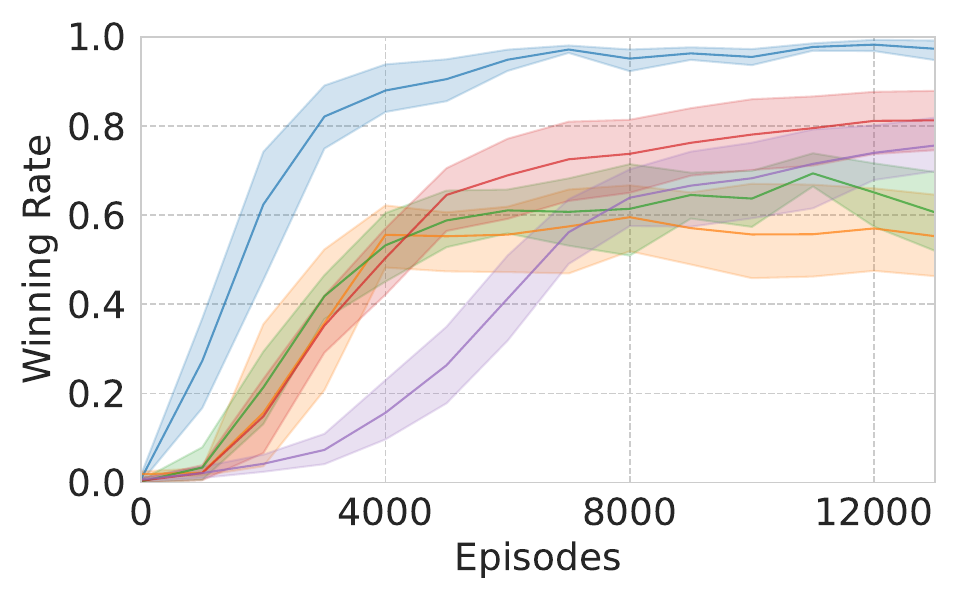}
		}
		\subfigure[MPE]{
			\label{fig:mpe_bot}
			\includegraphics[width=0.3\linewidth]{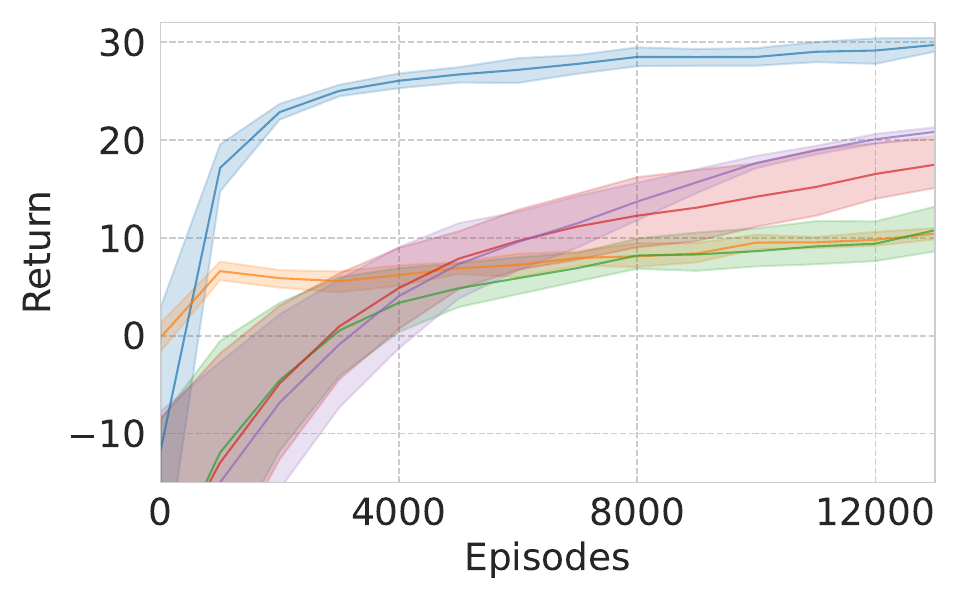}
		}
		\subfigure[RoboMaster]{
			\label{fig:rm_bot}
			\includegraphics[width=0.3\linewidth]{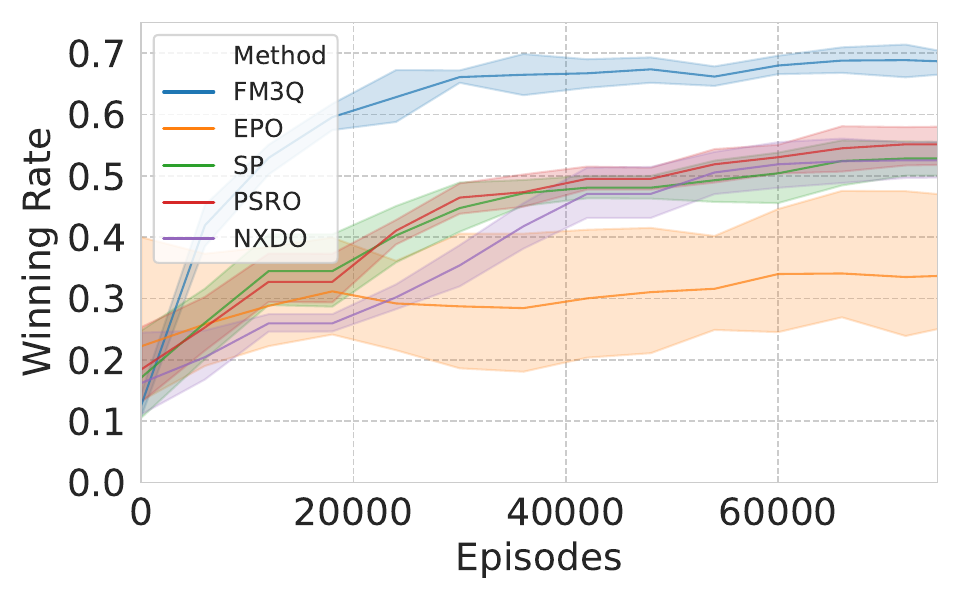}
		}
	\vspace{-2mm}
		\caption{Illustration of the performance of FM3Q during training by evaluating against the rule-based bots. }
		\label{fig:bot}
		
	\end{center}
\end{figure*}

\begin{figure*}[!ht]
	
	\begin{center}

		\setcounter{subfigure}{0}
		\subfigure[Wimblepong]{
			\label{fig:rr_pong}
			\includegraphics[width=0.3\linewidth]{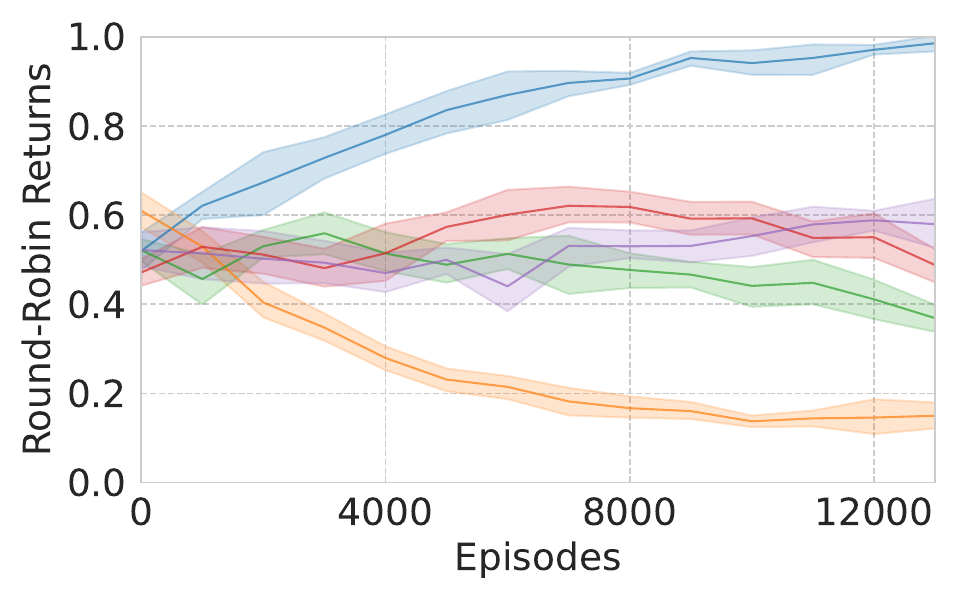}
		} \vspace{-3mm}
		\subfigure[MPE]{
			\label{fig:rr_mpe}
			\includegraphics[width=0.3\linewidth]{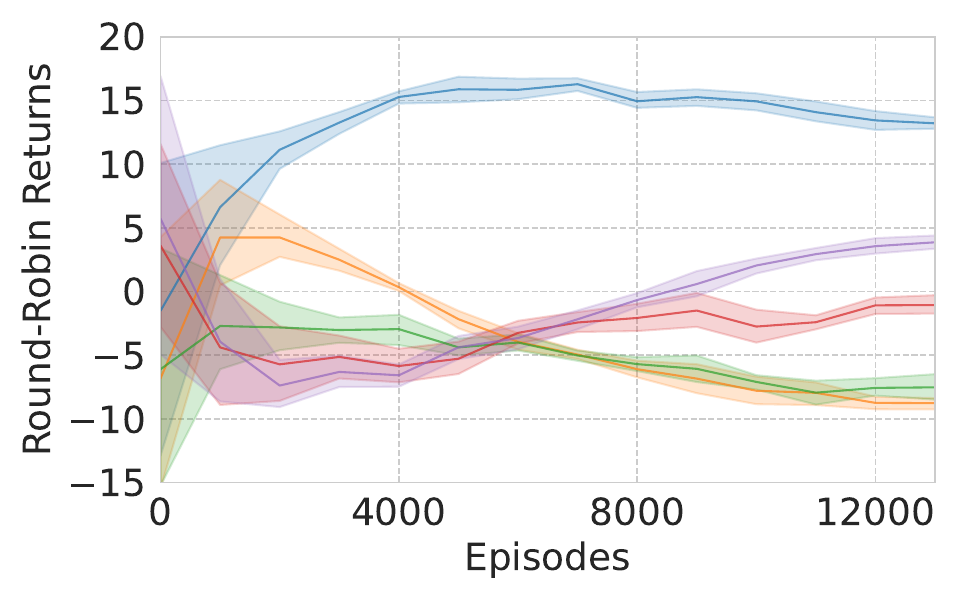}
		}
		\subfigure[RoboMaster]{
			\label{fig:rr_rm}
			\includegraphics[width=0.3\linewidth]{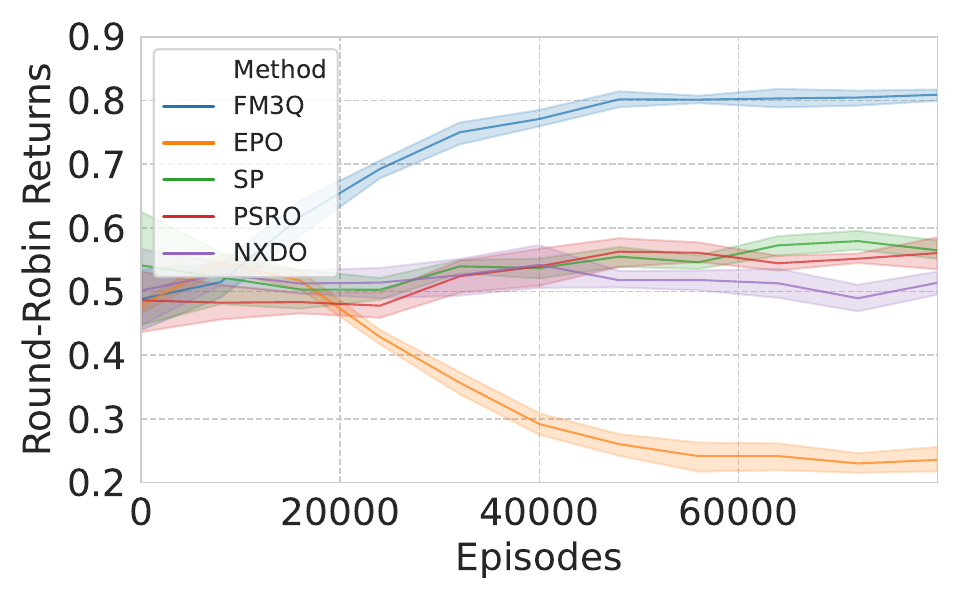}
		}
	
		\subfigure[Wimblepong]{
			\label{fig:bar_pong}
			\includegraphics[width=0.3\linewidth]{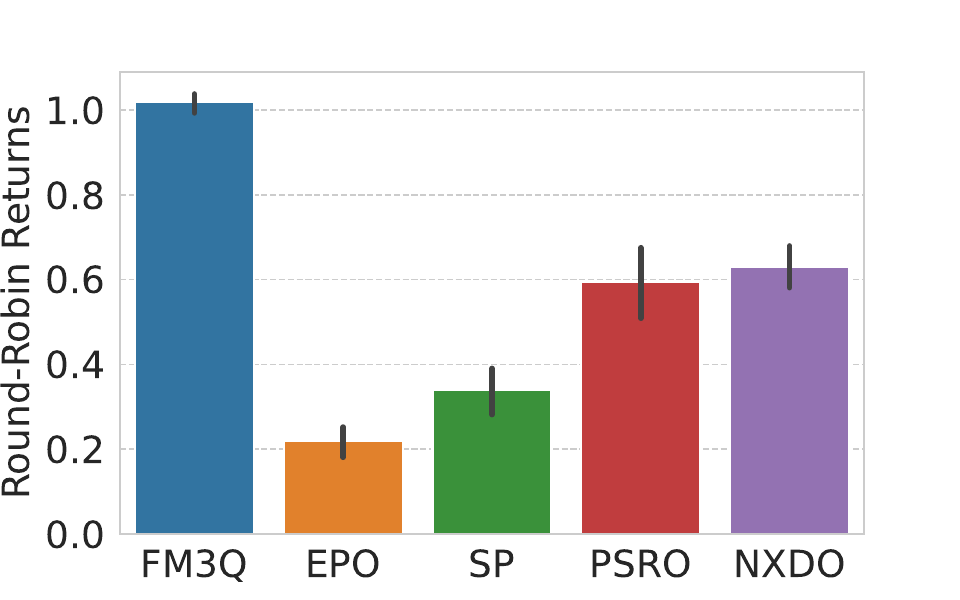}
		}
		\subfigure[MPE]{
			\label{fig:bar_mpe}
			\includegraphics[width=0.3\linewidth]{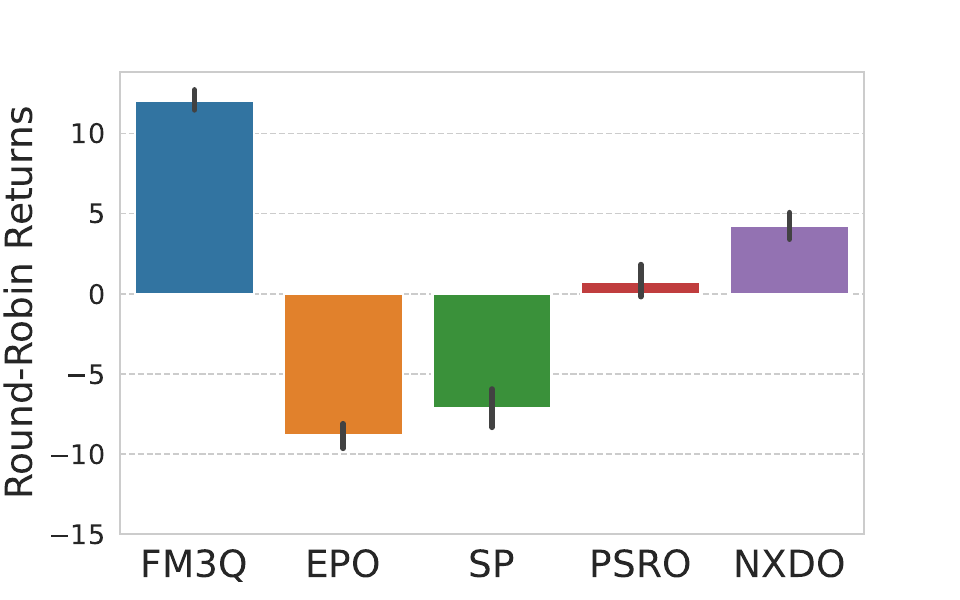}
		}
		\subfigure[RoboMaster]{
			\label{fig:bar_rm}
			\includegraphics[width=0.3\linewidth]{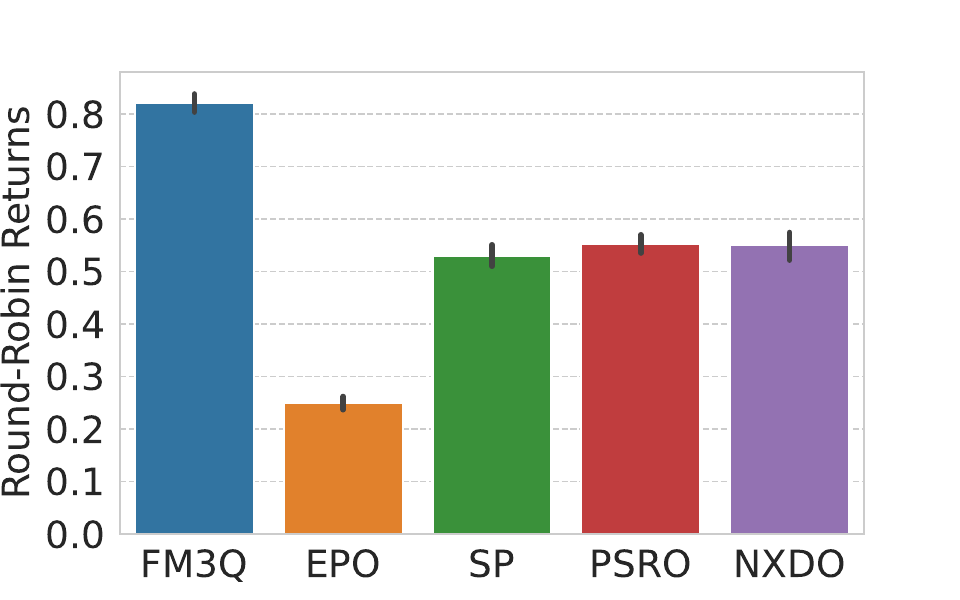}
		}\vspace{-2mm}
		\caption{Illustration of the Round-Robin results of FM3Q and baselines. (Top) normalised RR returns during training. (Bottom) normalised RR returns at the end of training.}
		
		\label{fig:rr}
		
	\end{center}
\end{figure*}

\begin{figure*}[!ht]
	
	\begin{center}

		\setcounter{subfigure}{0}
		\subfigure[Wimblepong]{
			\label{fig:nash_pong}
			\includegraphics[width=0.3\linewidth]{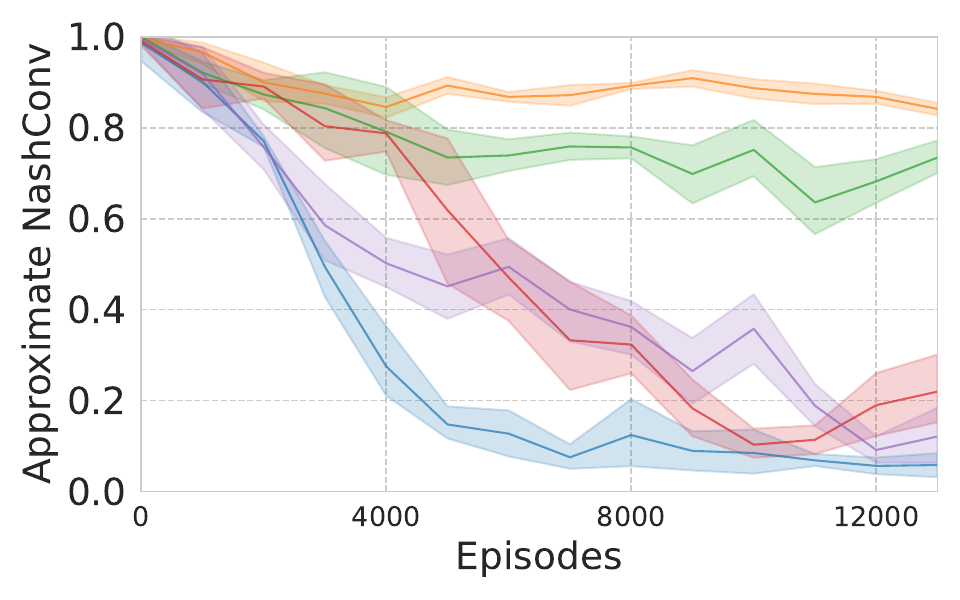}
		} 
		\subfigure[MPE]{
			\label{fig:nash_mpe}
			\includegraphics[width=0.3\linewidth]{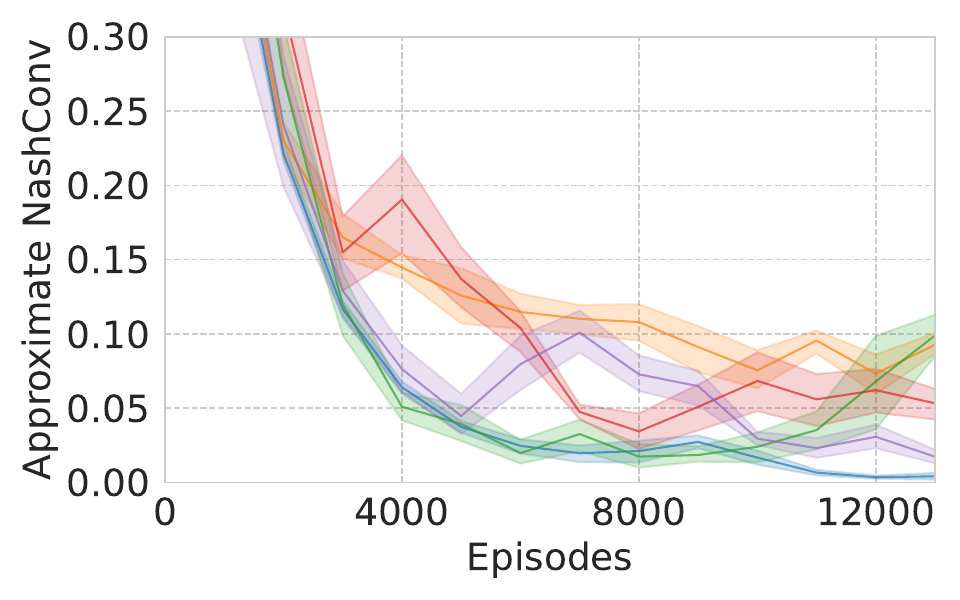}
		}
		\subfigure[RoboMaster]{
			\label{fig:nash_rm}
			\includegraphics[width=0.3\linewidth]{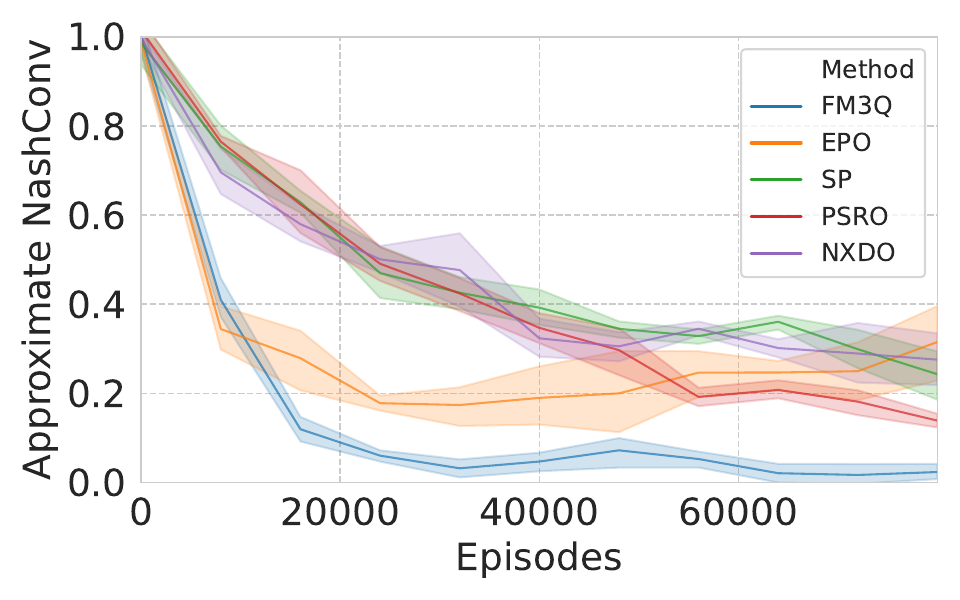}
		}
		\vspace{-2mm}
		\caption{Illustration of the approximate NashConv of FM3Q and baselines during training.}
		
		\label{fig:nash}
		
	\end{center}
\end{figure*}

\subsection {Performance against Script-based Bots}\label{sec:bot}

In this section, we utilize script-based bots to compare the performance differences between the three methods and the bots using the same amount of training data.
\textcolor{black}{In Pong and MPE, the script-based bots select actions to approach the ball, while in RoboMaster, the bot prioritizes attacking the unit with the least health.}
The model with superior performance should achieve a higher positive difference.
For the Wimblepong, MPE, and RoboMaster, we use a total of 13k, 13k, and 80k episodes to train FM3Q and baselines, and test the performance with the rule-based bots every 1k, 1k, and 8k episodes, respectively. The results of the learned models of all methods during their training process playing against the bots are shown in Figure \ref{fig:bot}. 

\textcolor{black}{
We can draw evident conclusions from the three result graphs: FM3Q exhibits the fastest optimization, the highest performance, and the smallest variance when compared to other baselines. Firstly, in Wimblepong, MPE, and RoboMaster, FM3Q requires fewer than 2k, 1k, and 10k episodes, respectively, to achieve performance of 0.5, 10, and 0.5. In contrast, the baselines necessitates a minimum of 4k, 6k, and 40k episodes, respectively, to reach the same performance levels in the three environments. It demonstrates the advantage of FM3Q in achieving better performance with less data.}

\textcolor{black}{
In addition, it is evident from the results that FM3Q outperforms other algorithms significantly. The performance of the models obtained by FM3Q in the three different environments against the bots is approximately 0.95, 30, and 0.7, respectively. In contrast, the best performance achieved by other algorithms in the three environments is as follows: PSRO achieves approximately 0.8 in Wimblepong, NXDO achieves approximately 20 in MPE, and NXDO achieves approximately 0.55 in RoboMaster.}

\textcolor{black}{
We observe that the performance fluctuations is minimal for FM3Q, followed by NXDO, PSRO, and SP, with the largest fluctuations observed for EPO. The primary reason for this is that FM3Q retains and effectively leverages all historical experience, while NXDO and PSRO engage more opponents relative to SP, making them relatively more stable. The reason for the inferior performance and variance of EPO lies in its single-agent optimization method, which lacks consideration for credit assignment among agents.}

\vspace{-1mm}

\subsection {Exploitability Evaluation}\label{sec:baseline}
\textcolor{black}{
The above experiments only demonstrate that, when competing with the fixed bots, FM3Q outperforms the baselines during the training process. Next, we utilize the different metrics, round-robin (RR) tournament results to estimate the exploitability of each method against all other baselines \cite{samvelyan2022maestro}, and approximate NashConv when facing the best response \cite{timbers2022approximate}.}

\textcolor{black}{
We evaluate RR tournament results where each method is matched against every other method. A round-robin tournament is a competition where each contestant meets every other participant. In particular, during the training process, we preserve models of all methods across different training phases. Then we select models from each method at the same training phase and engage them in cross-play to obtain outcomes. Finally, the method accumulates its scores against all other methods. Less exploitable agents should attain a higher RR returns than all other agents.
The three subplots at the top of Figure \ref{fig:rr} respectively depict the normalised RR returns of all methods during training in three environments, and
show that FM3Q is able to quickly improve its performance and outperforms baselines.
The three subplots at the bottom of Figure \ref{fig:rr} display the normalised RR returns at the end of training, and show that FM3Q outperforms all baselines and highlight the superiority of FM3Q.}

\textcolor{black}{
Next, we employ the NashConv metric to demonstrate the efficacy of the policy. Calculating NashConv necessitates finding the best response to the policy, yet precise best responses are unattainable in the current experimental setting. Therefore, we resort to using an approximate NashConv. We train the RL-based agents independently, allowing them to compete against periodically stored models for evaluating the performance improvements achievable by these RL-based agents. Less exploitable agents should attain a lower approximate NashConv than all other agents.
The approximate Nashconv curves of models trained in three different environments are depicted in Figure \ref{fig:nash}. It can be observed that, across these three environments, the approximate Nashconv of FM3Q consistently exhibits a decreasing trend throughout the training process, eventually converging to very small values. Furthermore, it outperforms the contemporaneous models from other algorithms, indicating that FM3Q possesses superior exploitability. NXDO and PSRO also demonstrate decreasing trends, with their values trailing behind FM3Q but surpassing those of SP and EPO during the same time period. Notably, the Nashconv curve of SP exhibits significant fluctuations in the Pong and MPE environments, while EPO attains the highest level of exploitability.}

\begin{figure*}[!ht]
	
	\begin{center}
		\subfigure[Small buffer size on Wimblepong.]{
			\label{fig:pong_s}
			\includegraphics[width=0.25\linewidth]{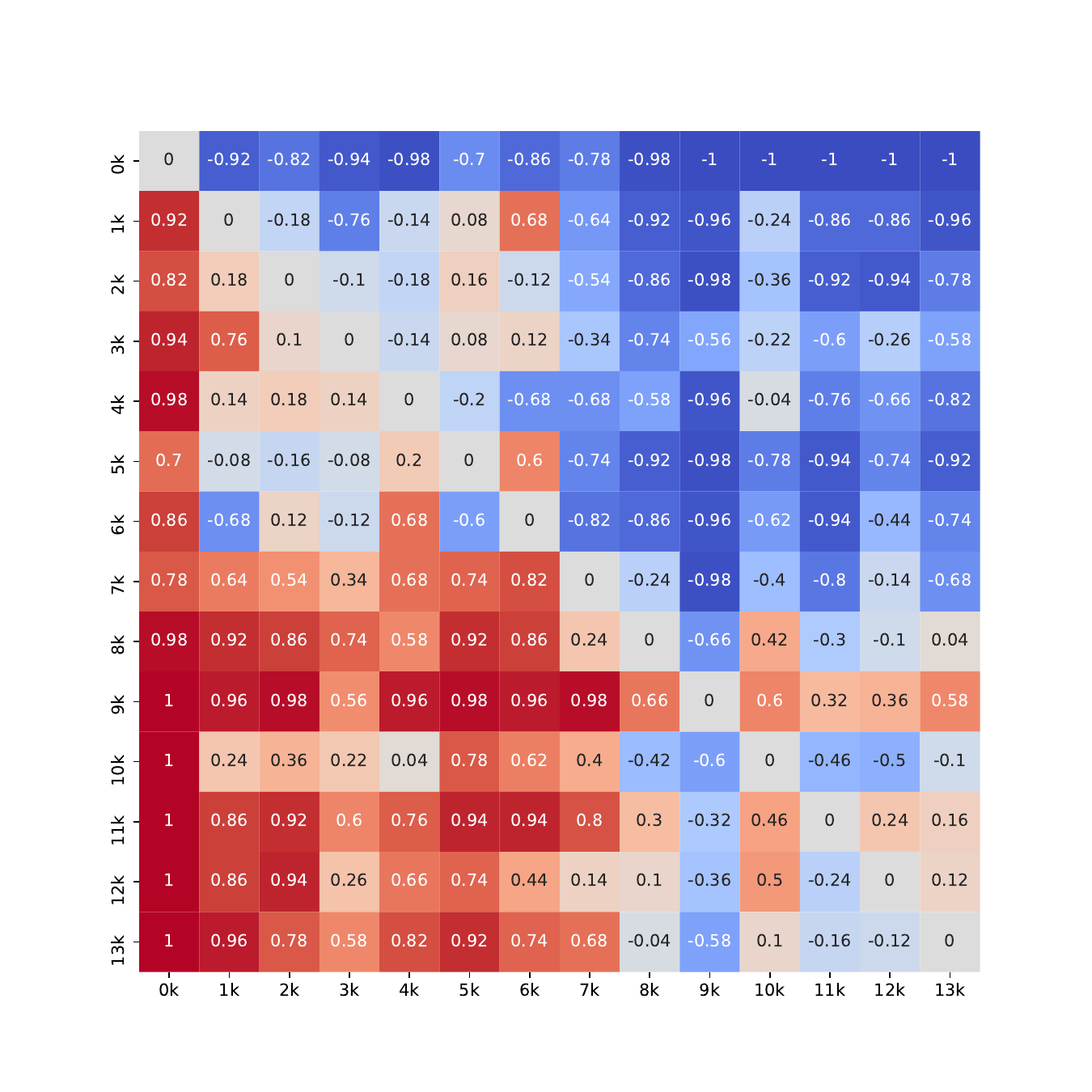}
		}
		\subfigure[Large buffer size on Wimblepong.]{
			\label{fig:pong_l}
			\includegraphics[width=0.25\linewidth]{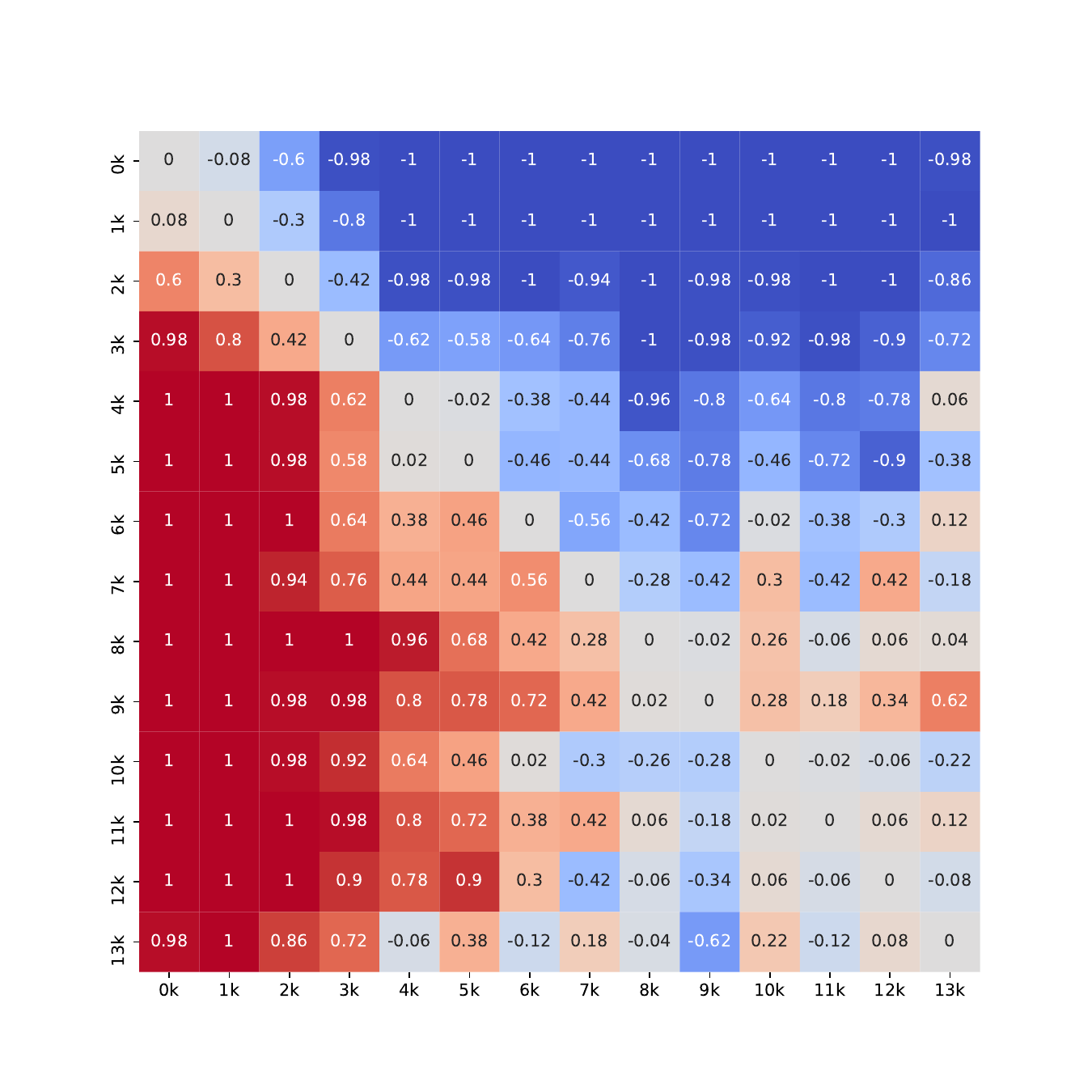}
		}
		\subfigure[Full buffer size on Wimblepong.]{
			\label{fig:pong_f}
			\includegraphics[width=0.3\linewidth]{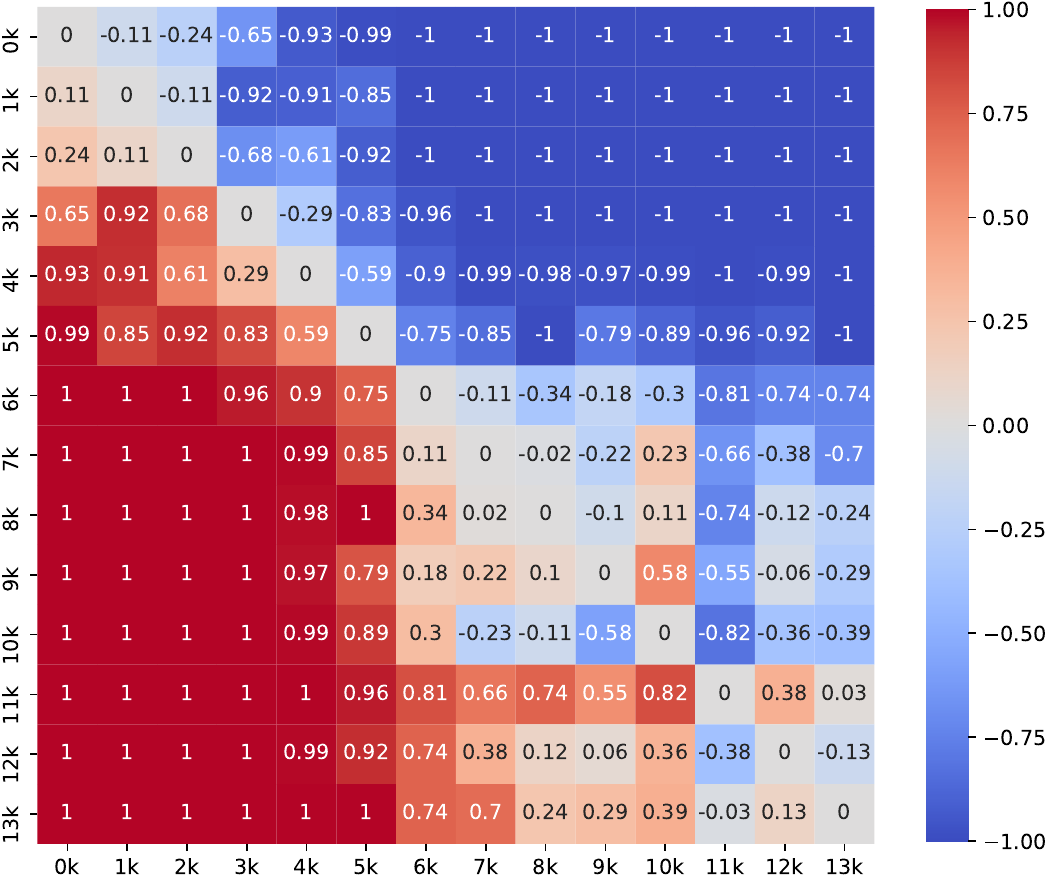}
		}
	\end{center}
	\vspace{-7mm}
	\begin{center}
		\subfigure[Small buffer size on MPE.]{
			\label{fig:mpe_s}
			\includegraphics[width=0.25\linewidth]{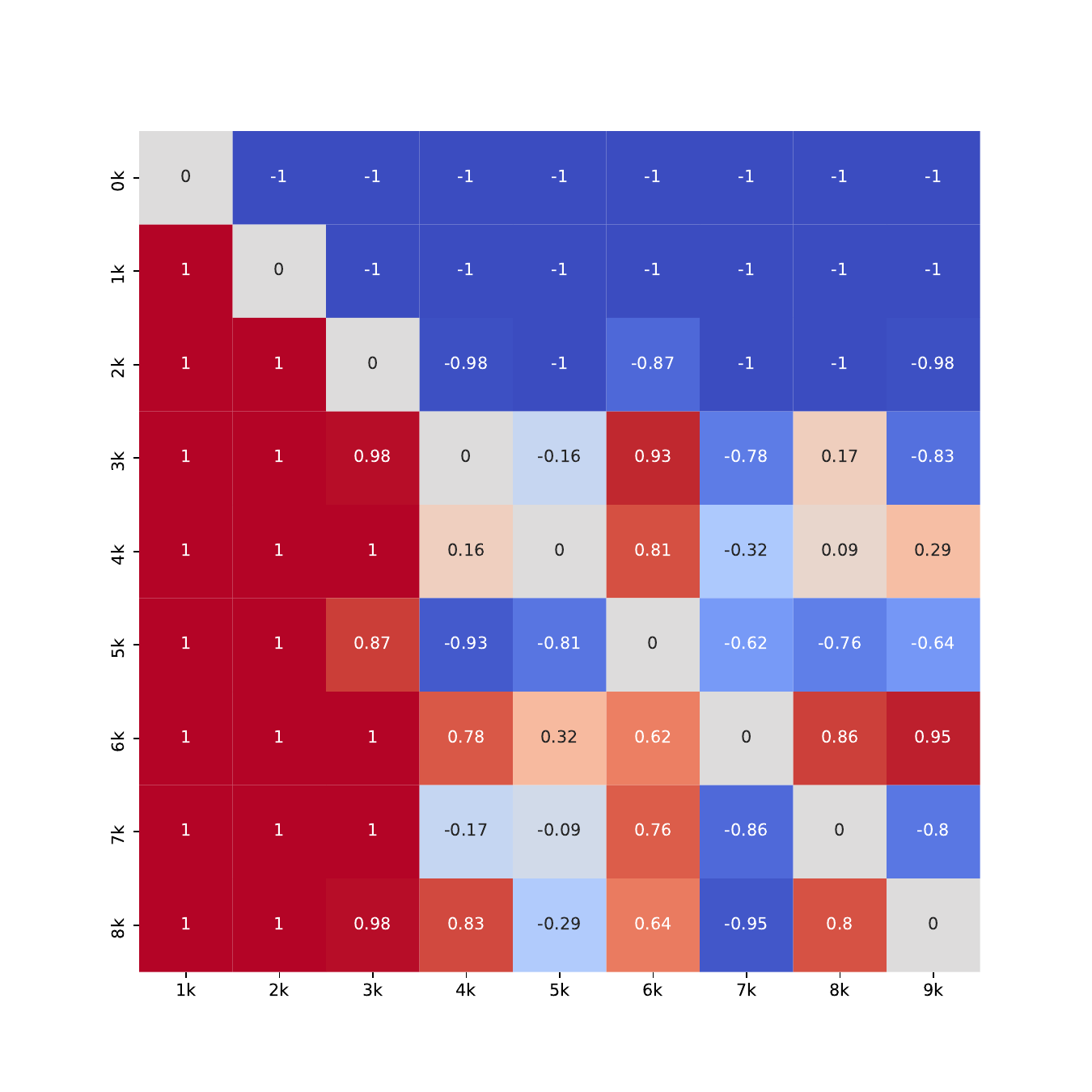}
		}
		\subfigure[Large buffer size on MPE.]{
			\label{fig:mpe_l}
			\includegraphics[width=0.25\linewidth]{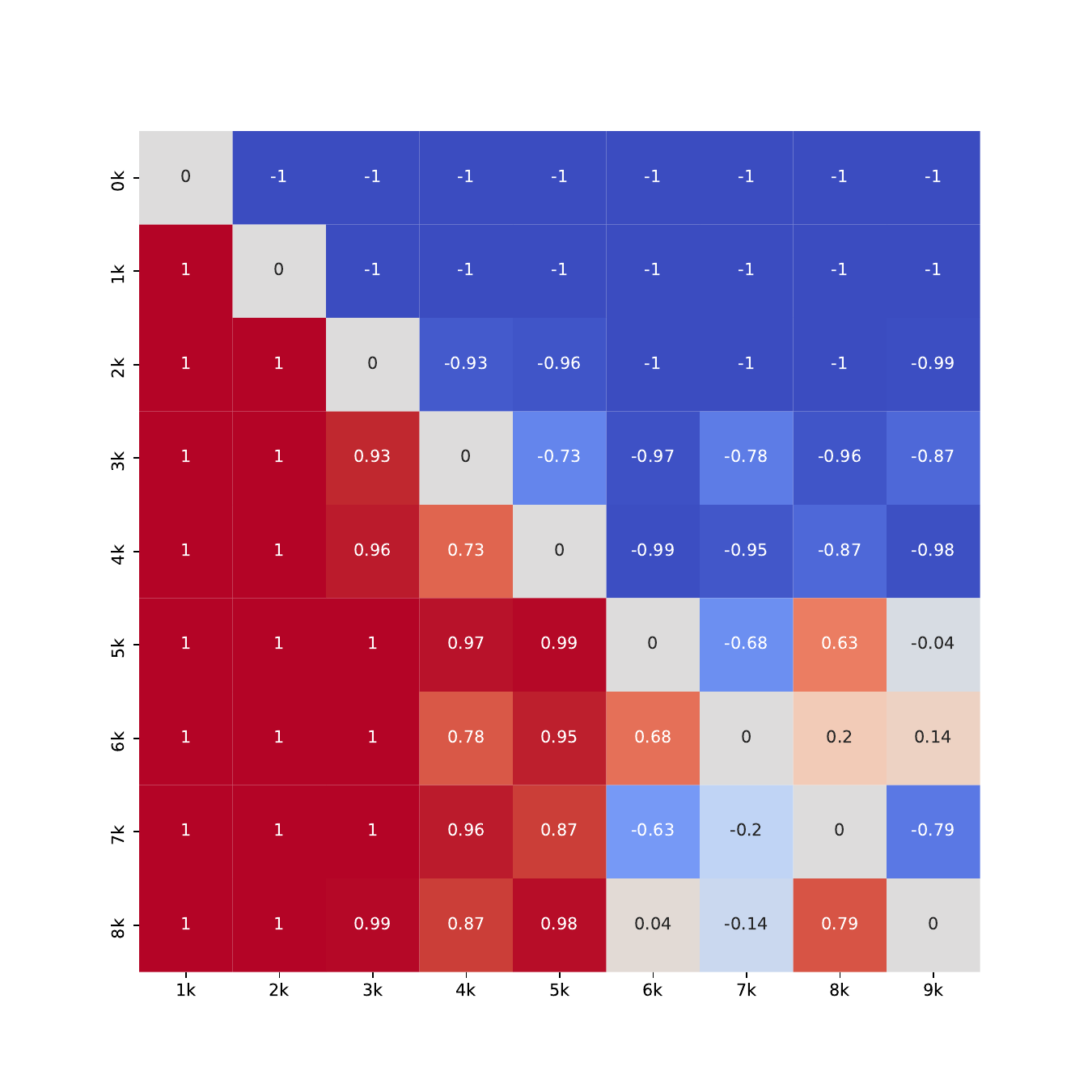}
		}
		\subfigure[Full buffer size on MPE.]{
			\label{fig:mpe_f}
			\includegraphics[width=0.3\linewidth]{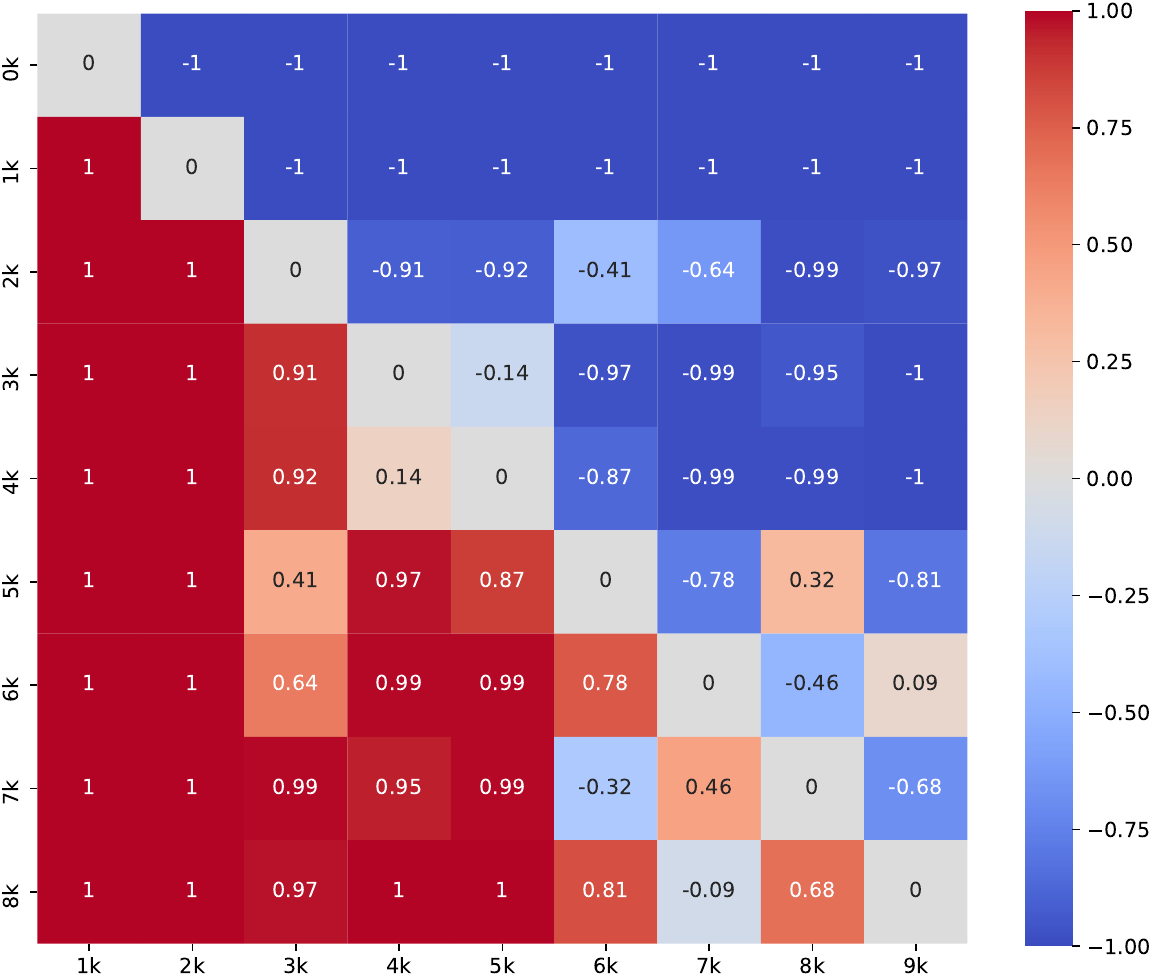}
		}
	\end{center}
	\vspace{-7mm}
	\begin{center}
		\subfigure[Small buffer size on RoboMaster.]{
			\label{fig:rm_s}
			\includegraphics[width=0.25\linewidth]{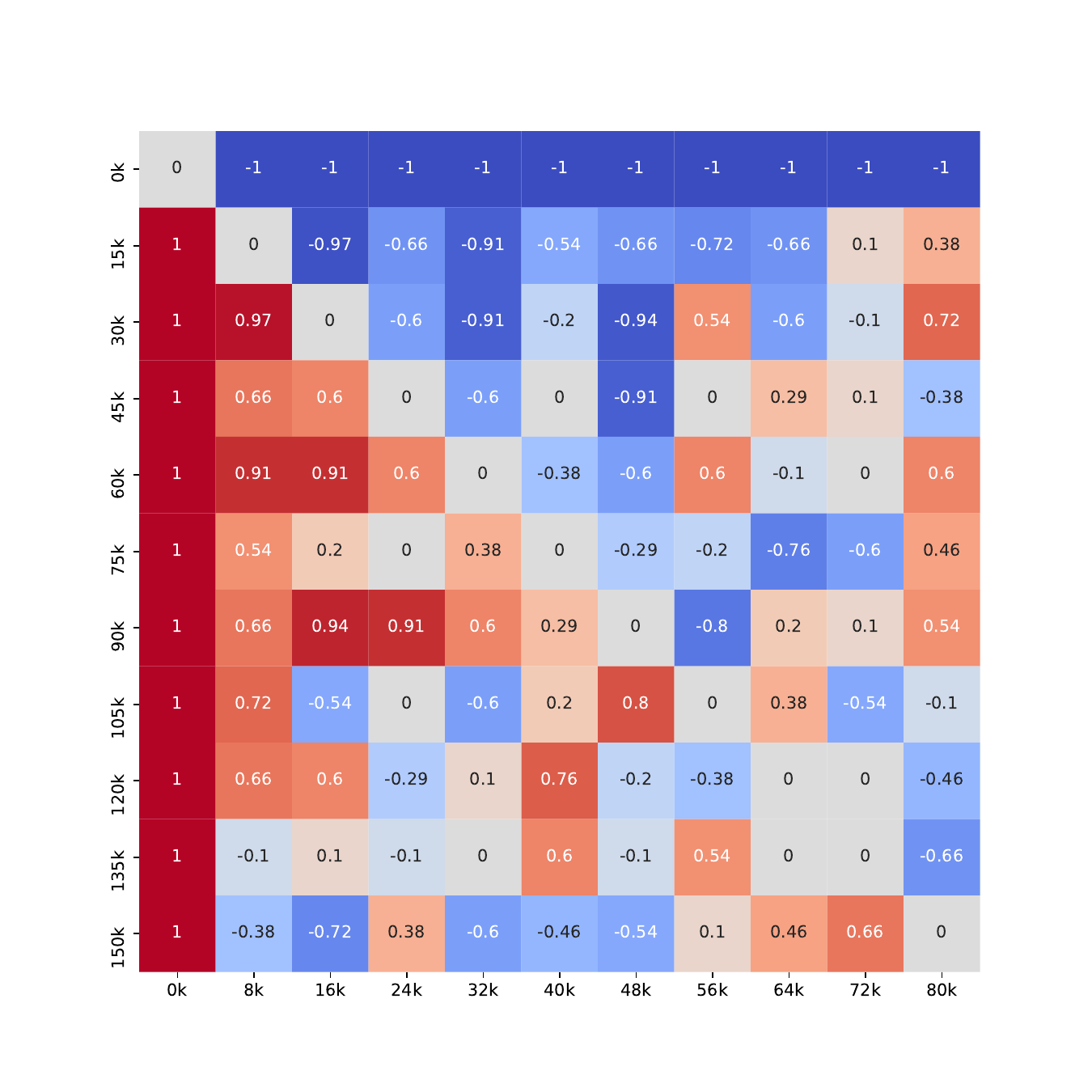}
		}
		\subfigure[Large buffer size on RoboMaster.]{
			\label{fig:rm_l}
			\includegraphics[width=0.25\linewidth]{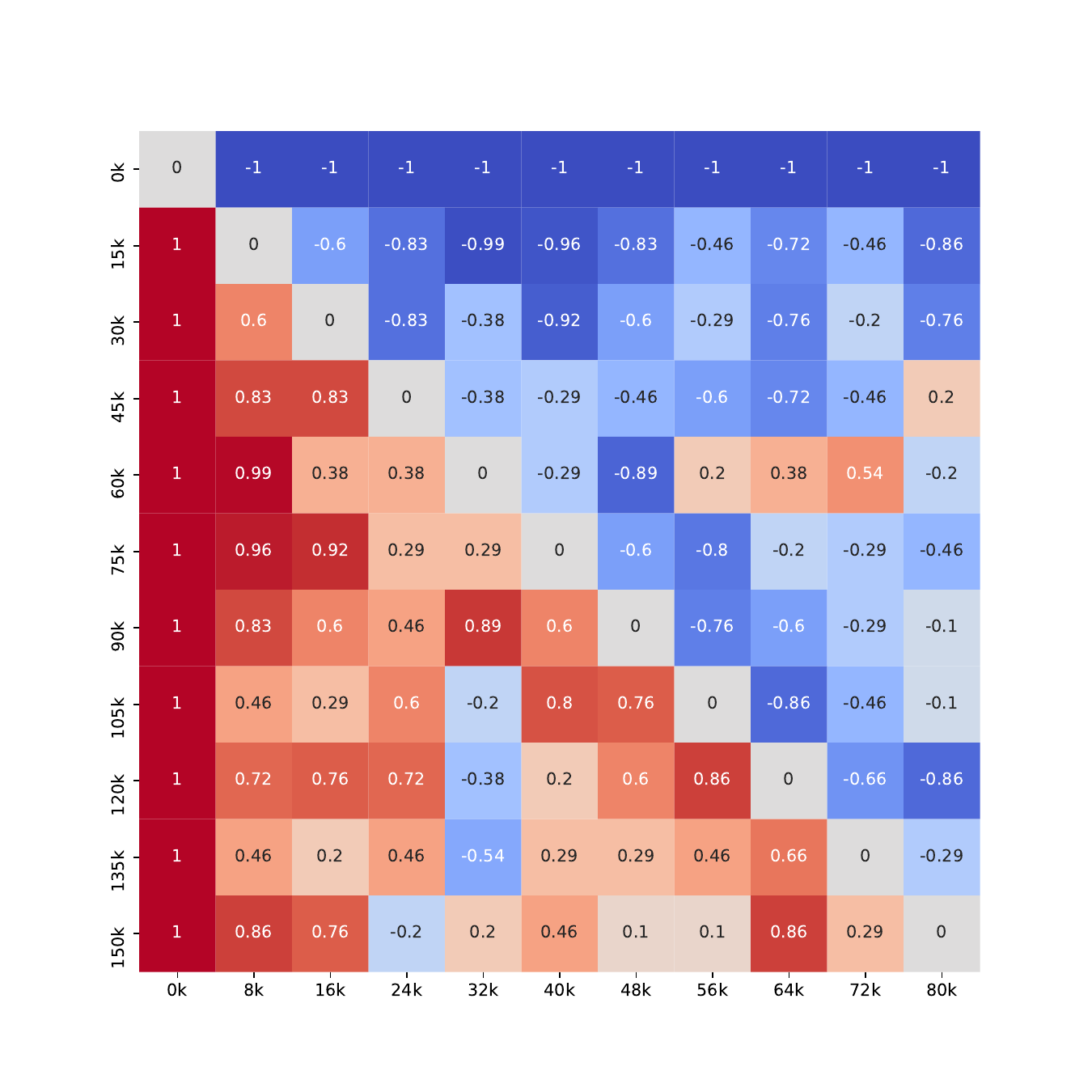}
		}
		\subfigure[Full buffer size on RoboMaster.]{
			\label{fig:rm_f}
			\includegraphics[width=0.3\linewidth]{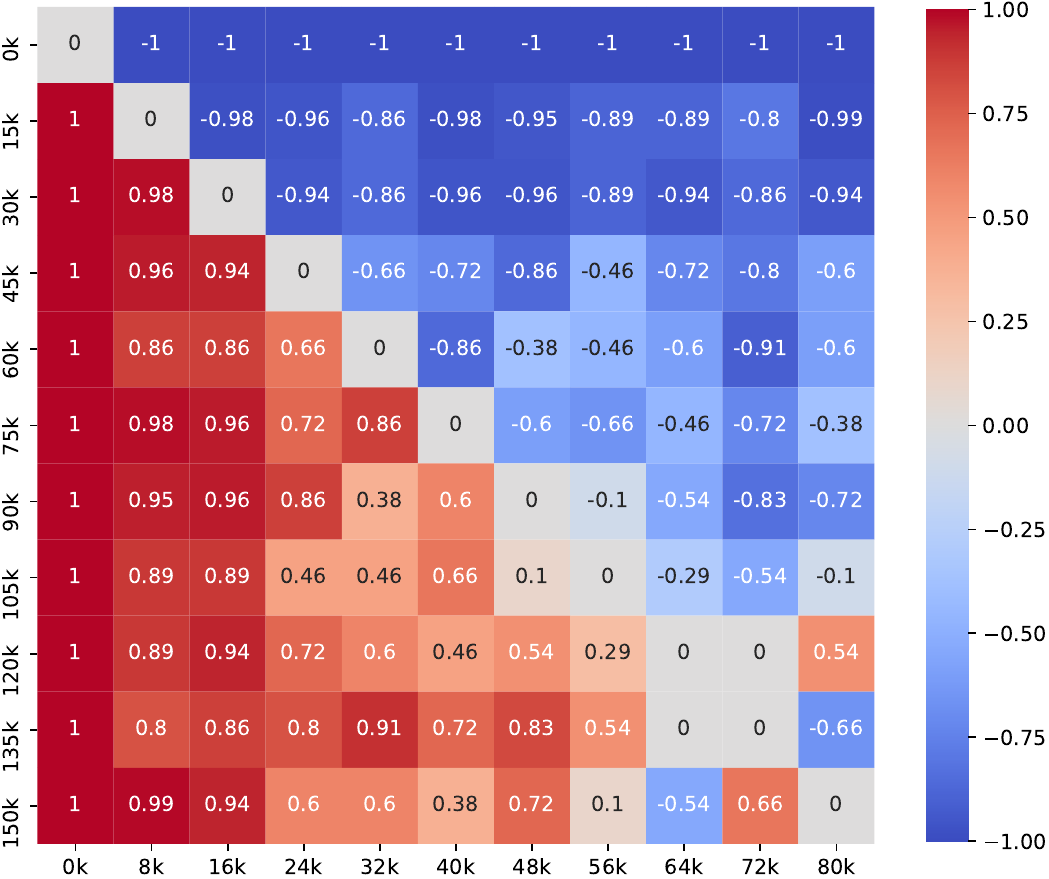}
		}
	\end{center}
	\vspace{-4mm}
	\caption{\textcolor{black}{Illustration of the payoff values of FM3Q populations in the three scenarios under different replay buffer sizes. In each scenario, from left to right, the red squares on the lower triangle portion of each table become darker as the buffer size gradually increases. In other words, the proportion of new policies that outperform the old policies increases, and the degree of outperformance also becomes higher.}}
	\label{fig:ab}
\end{figure*}

\begin{figure*}[!ht]
	
	\begin{center}

		\setcounter{subfigure}{0}
		\subfigure[Wimblepong]{
			\label{fig:min_ab_pong}
			\includegraphics[width=0.3\linewidth]{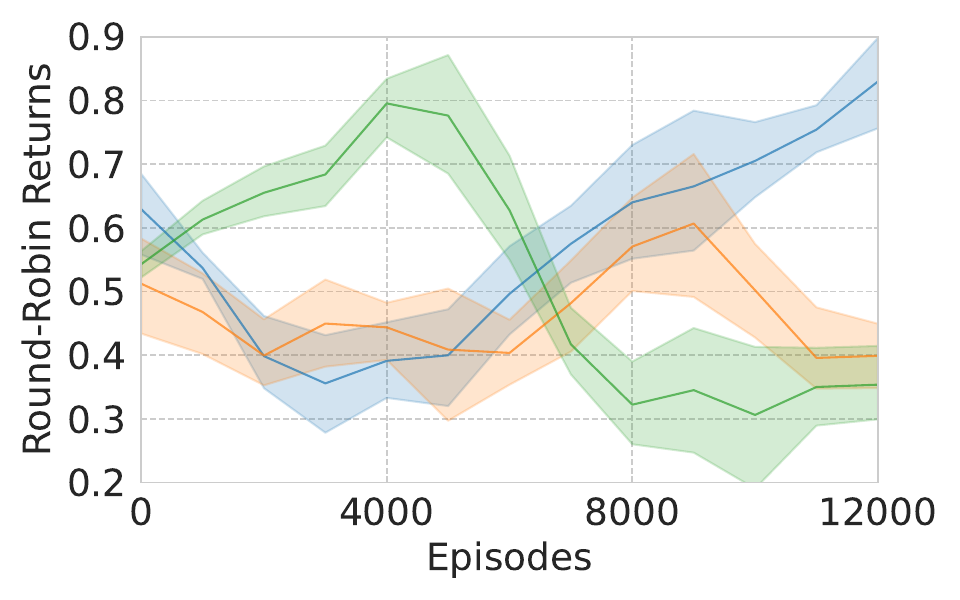}
		}
		\subfigure[MPE]{
			\label{fig:min_ab_mpe}
			\includegraphics[width=0.3\linewidth]{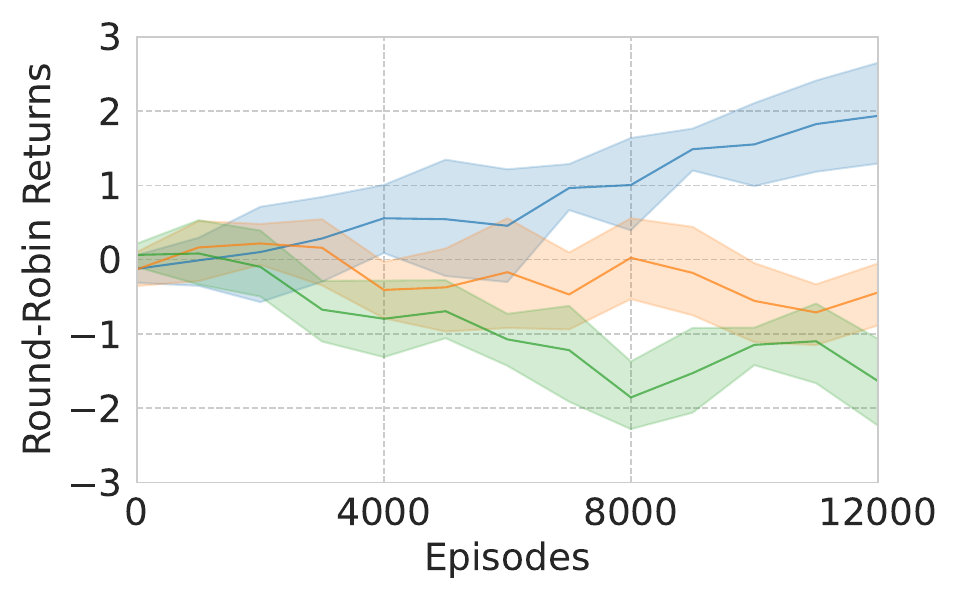}
		}
		\subfigure[RoboMaster]{
			\label{fig:min_ab_rm}
			\includegraphics[width=0.3\linewidth]{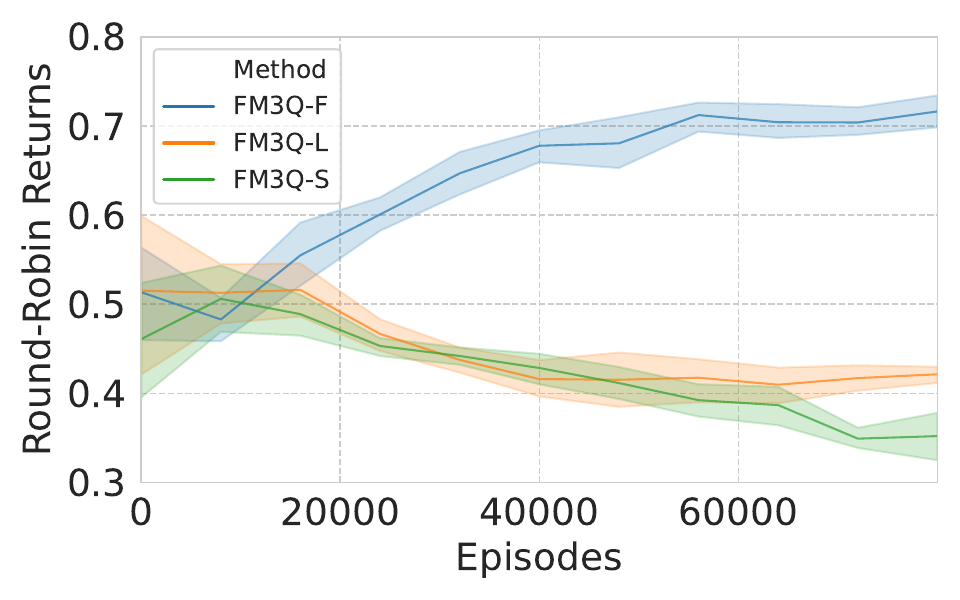}
		}
		\vspace{-2mm}
		\caption{Illustration of the Round-Robin returns of FM3Q-S, FM3Q-L, and FM3Q-F during training.}
		
		\label{fig:rr_ab}
		
	\end{center}
\end{figure*}

\vspace{-1mm}
\subsection {Optimization Trend of FM3Q}\label{sec:process}
We then investigate the performance of FM3Q across different training phases and display the results using payoff values calculated using the average winning rates, or returns by cross-play. 
Figure \ref{fig:pong_f}, Figure \ref{fig:mpe_f}, and Figure \ref{fig:rm_f} show the payoff tables for Wimblepong, MPE, and RoboMaster, respectively. 
Taking Figure \ref{fig:pong_f} as an example, we save a total of 14 models at different phases. These models are evaluated using the round-robin tournament results, and the average winning rate between each pair of models is presented in the form of a payoff value.
The red squares indicate victories, with darker shades representing higher winning rates. Conversely, the blue squares indicate defeats, with darker shades representing lower winning rates.

It can be seen that the squares in the lower triangle of the tables are almost all red, and that the squares in the same row always have darker shading on the left than on the right. So, the first thing we can say is that the models that are trained later can almost always do better than the ones that are trained earlier.

Of course, in the lower triangle of each of the three payoff tables, there are a few individual cells that are colored blue. However, the proportion of these cells is very small, meaning that the probability of the models trained later being defeated by the earlier ones is extremely low. In the lower triangle of Figure \ref{fig:pong_f}, Figure \ref{fig:mpe_f}, and Figure \ref{fig:rm_f}, these cells only account for $5/196$, $2/64$, and $1/121$, respectively. It can be concluded from this phenomenon that, the later the model is learned, the better it will perform. 

\subsection {Ablation Study}\label{sec:ab}
We argue that using a larger buffer as much as possible is conducive to training FM3Q; we even need to use all the data generated by interaction in some tasks. We investigate the effect of historical experience in FM3Q and demonstrate the above argument by running experiments with different sizes of replay buffer. For the purpose of distinguishing among different replay buffer sizes, we refer to the FM3Q trained on small, large, and full buffer sizes as FM3Q-S, FM3Q-L, and FM3Q-F, respectively.

First, we investigate the effect of historical experience in the optimization tread of FM3Q. We compare their performance across different training phases and display the results using payoff values calculated using the average winning rates, or returns. Figure \ref{fig:ab} shows the learning performance of FM3Q-S, FM3Q-L, and FM3Q-F in different replay buffer size sets in the three scenarios. The replay buffer size from left to right in each scenario (each row) decreases in turn. We focus on the lower triangle portion of each table and find that, as replay buffer size gradually increases, the red part becomes more and more dark, while the blue part becomes less and lighter. It is concluded that, as the buffer size increases, the proportion of new policies that outperform the old policies increases, and the degree of outperformance also becomes higher. A small replay buffer causes instability in FM3Q’s learning. It is worth noting that a complete or sufficient replay of historical experience causes FM3Q to update policies in a transitive or monotone mode. Therefore, the results are consistent with the argument.

\textcolor{black}{
Second, we also utilize the RR returns as comparative metrics to estimate the exploitability of FM3Q with each buffer size against all other sizes.
We evaluate all pairs of methods in cross-play during their training processes. Less exploitable agents should attain a higher RR returns than all other agents.
Figure \ref{fig:rr_ab} displays the performance of FM3Q-S, FM3Q-L, and FM3Q-F on the three scenarios. We can see that FM3Q-F finally outperforms FM3Q-S and FM3Q-L in RR returns. Therefore, we can conclude that an increase in replay buffer size contributes to the improvement of the training effectiveness and performance of the FM3Q.}

\section {Conclusion}\label{sec:Conclusion}

To design an efficient MARL framework for 2t0sMGs, we propose the IGMM to extend factorizable tasks to 2t0sMGs. Leveraging the IGMM principle, we introduce FM3Q and its online learning algorithm based on CTDE to factorize the joint minimax Q function into individual ones while synchronously optimizing the networks of all agents. Additionally, we prove the convergence of FM3Q and empirically demonstrate the superiority of FM3Q over existing methods in terms of learning efficiency and final performance. 
However, there is scope for improvement in the FM3Q algorithm. First of all, because FM3Q's policy is deterministic, it can be challenging or impossible to identify the exact NE in some competitive tasks that only use mixed (stochastic) policies. Secondly, FM3Q retains historical experience, requiring all experience to be trained before updating the target network. This results in increased hardware costs for storage and computation, particularly in large games. We suggest incorporating stochastic policy methods into FM3Q to ensure convergence to the NE. In addition, the optimization efficiency can be improved by performing quality filtering on the data in the buffer to remove low-quality samples.

\section {Appendix}\label{sec:Appendix}

For the sake of fairness, we utilize QMIX as an individual in the SP, PSRO and NXDO policy populations. FM3Q, SP, PSRO, and NXDO are realized in the framework of QMIX implementation in \href{https://github.com/starry-sky6688/MARL-Algorithms}{https://github.com/starry-sky6688/MARL-Algorithms}. The important hyperparameters of all methods in experiments are listed in Table \ref{tab:hp}. n\_ep, mix\_hidden\_dim, n\_gene, and ep\_per\_gene represent the number of training episodes, the dimension of the hidden layers in the mixing networks, the number of generations, and the number of training episodes per generation, respectively.

\vspace{-2mm}
\begin{table}[!h]
	{\fontsize{8}{12}\selectfont
	\centering
	\caption{\textcolor{black}{The important hyperparameters of all methods in Experiments.}}
	\begin{tabular}{|c|c|c|c|c|}
		\hline
		\hline
		Alg. & Hyperparameters & Wimblepong & MPE  & RoboMaster \\
		\hline
		\multirow{6}{*}{\textbf{Common}} &	n\_episodes & 1.3e4 & 1.3e4 &  8e4 \\
		& n\_seeds & 8 & 8 &  8 \\
		& gamma & 0.99 & 0.98 &  0.99 \\
		& hidden\_layers & [64,64] & [64,64]  &  [128,128]  \\
		& mix\_hidden\_dim & 32 & 32 &  32 \\
		& learning\_rate & 5e-4 & 5e-4 &  5e-4 \\
		\hline
		\multirow{3}{*}{\textbf{FM3Q}} & full buffer\_size & 4e6 & 4e5&  4e6 \\
		 &	large buffer\_size & 1e6 & 8e4 &  1e6 \\
		& small buffer\_size & 2e5 & 2e4 &  4e5 \\
		\hline
	    \multirow{4}{*}{\parbox{1cm}{\centering \textbf{SP/ \\ PSRO/ \\ NXDO}}} &	n\_genes & 13 & 13 &  10 \\
		 &ep\_per\_gene & 1e3 & 1e3 &  8e4 \\
		 &batch\_size & 1e3 & 1e3 &  2e3 \\
		 &buffer\_size & 2e5 & 2e4 &  2e5 \\
		\hline
		\multirow{4}{*}{\textbf{EPO}} & pi\_learning\_rate & 3e-4 & 3e-4 & 3e-4 \\
		&vf\_learning\_rate & 1e-3 & 1e-3 & 1e-3 \\
		&train\_iters & 80 & 80 & 80 \\
		&target\_kl & 0.02 & 0.02 & 0.02 \\
		\hline
	\end{tabular}%
	\label{tab:hp}%
}
\end{table}%

\bibliographystyle{IEEEtran}
\small
\bibliography{ref}
%






\end{sloppypar}

\end{document}